\newif\ifarxiv

\arxivfalse
\arxivtrue

\ifarxiv
\documentclass[sigconf,10pt,nonacm=true,anonymous=false]{acmart}
\else
\documentclass[sigconf,10pt,nonacm=false,anonymous=false]{acmart}
\fi
\AtBeginDocument{%
  }

\setcopyright{acmcopyright}
\copyrightyear{2024}
\acmYear{2024}
\acmDOI{XXXXXXX.XXXXXXX}

\acmConference[MobiCom '24]{Make sure to enter the correct
  conference title from your rights confirmation emai}{November 18--22,
  2024}{Washington DC}
\acmPrice{15.00}
\acmISBN{978-1-4503-XXXX-X/18/06}

\usepackage{algorithmic}
\usepackage{adjustbox}
\usepackage{graphicx}
\usepackage{textcomp}
\usepackage{xcolor}
\usepackage{soul}
\usepackage{booktabs}
\usepackage{multirow}
\usepackage{subcaption}
\usepackage{xspace}
\usepackage[ruled,linesnumbered,algo2e,noend]{algorithm2e}
\usepackage{enumitem}
\usepackage{gensymb}

\newif\ifcomment
\commentfalse

\ifcomment
\newcommand{\steve}[1]{\sethlcolor{cyan}\hl{[Steve: #1]}}
\newcommand{\minos}[1]{\sethlcolor{orange}\hl{[Minos: #1]}}
\newcommand{\lorenzo}[1]{\sethlcolor{green}\hl{[Lorenzo: #1]}}
\newcommand{\hamed}[1]{\sethlcolor{yellow}\hl{[Hamed: #1]}}
\newcommand{\camready}[1]{{#1}}
\else
\newcommand{\steve}[1]{}
\newcommand{\minos}[1]{}
\newcommand{\lorenzo}[1]{}
\newcommand{\hamed}[1]{}
\newcommand{\camready}[1]{{#1}}
\fi

\ifarxiv
\usepackage{background}
\backgroundsetup{angle=0, 
scale=1,
color=black,
firstpage=true,
position=current page.north,
hshift=0pt,
vshift=-20pt,
contents={\ifnum\value{page}=1 PREPRINT: Accepted at the 30th Annual International Conference On
Mobile Computing And Networking (MobiCom'24) \else \fi}
}
\fi

\newcommand{\tool}{{\sc MELT}\xspace}

\SetKwIF{If}{ElseIf}{Else}{if}{:}{elif}{else}{end if}
\SetKwFor{For}{for}{:}{endfor for}%

\let\oldnl\nl%
\newcommand{\nonl}{\renewcommand{\nl}{\let\nl\oldnl}}%

\begin{document}

\title{MELTing point: Mobile Evaluation of Language Transformers}

\author{\mbox{Stefanos Laskaridis}}
\ifarxiv
\fi
\email{mail@stefanos.cc}
\affiliation{%
  \institution{Brave Software}
  \city{London}
  \country{UK}
}

\author{Kleomenis Katevas}
\email{kkatevas@brave.com}
\affiliation{%
  \institution{Brave Software}
  \city{London}
  \country{UK}
}

\author{Lorenzo Minto}
\email{lminto@brave.com}
\affiliation{%
  \institution{Brave Software}
  \city{London}
  \country{UK}
}

\author{Hamed Haddadi}
\email{hhaddadi@brave.com}
\affiliation{%
  \institution{Brave Software \& Imperial College London}
  \city{London}
  \country{UK}
}

\renewcommand{\shortauthors}{Laskaridis et al.}

\begin{abstract}

Transformers have recently revolutionized the machine learning (ML) landscape, gradually making their way into everyday tasks and equipping our computers with ``sparks of intelligence''. However, their runtime requirements have prevented them from being broadly deployed on mobile. As personal devices become increasingly powerful at the consumer edge and prompt privacy becomes an ever more pressing issue, we explore the current state of mobile execution of Large Language Models (LLMs). To achieve this, we have created our own automation infrastructure, \tool, which supports the headless execution and benchmarking of LLMs on device, supporting different models, devices and frameworks, including Android, iOS and Nvidia Jetson devices. We evaluate popular instruction fine-tuned LLMs and leverage different frameworks to measure their end-to-end and granular performance, tracing their memory and energy requirements along the way. 

Our analysis is the first systematic study of on-device LLM execution, quantifying performance, energy efficiency and accuracy across various state-of-the-art models and showcases the state of on-device intelligence in the era of hyperscale models. Results highlight the performance heterogeneity across targets and corroborates that LLM inference is largely memory-bound. Quantization drastically reduces memory requirements and renders execution viable, but at a non-negligible accuracy cost. Drawing from its energy footprint and thermal behavior, the continuous execution of LLMs remains elusive, as both factors negatively affect user experience. Last, our experience shows that the ecosystem is still in its infancy, and algorithmic as well as hardware breakthroughs can significantly shift the execution cost. We expect NPU acceleration, and framework-hardware co-design to be the biggest bet towards efficient standalone execution, with the alternative of offloading tailored towards edge deployments.

\end{abstract}


\maketitle

\section{Introduction}
\label{sec:intro}

Our devices are getting increasingly more capable in performing tasks that have traditionally required human intelligence~\cite{bubeck2023sparks,schaeffer2024emergent}. The proliferation of capable on-device hardware has enhanced their capabilities in areas such as vision~\cite{radford2021learning,dosovitskiy2021an}, language~\cite{radford2019language,vernikos2023small} and sensor understanding~\cite{10.1145/3638550.3641130}. Convolutional~\citep{krizhevsky2012imagenet} and recurrent architectures~\citep{gers2000learning} have undoubtedly been fueling intelligence over the past decade, with significant investment towards their performant execution, via hardware~\citep{venieris2016fpgaconvnet}, algorithmic~\citep{leviathan2023fast,laskaridis2021adaptive} and EfficientML~\citep{frankle2018the,lin2023awq,hinton2015distilling} optimizations, along with hardware and software co-design~\citep{kouris2022fluid,abdelfattah2020best}.

Lately, however, transformers~\citep{vaswani2017attention} have become the {go-to} architecture for deep learning models, with attention mechanisms offering unparalleled performance and the ability to model long-sequence data with fewer inductive biases. Applied across modalities, including vision, speech and text~\cite{dosovitskiy2021an,radford2019language,radford2023robust}, these models have demonstrated significant performance benefits, across modeling and generation tasks. Another key benefit of these models has been their ability to scale to very large sizes, both in terms of data ingestion and parameter size, without their performance plateauing~\cite{touvron2023llama2}. This has given birth to ``foundation models'', large models that are trained on large corpora of data and act as universal backbones for a series of downstream tasks. For example, Large Language Models (LLMs)~\citep{touvron2023llama2,chung2022scaling,hoffmann2022training} have been trained on large parts of the Internet~\cite{commoncrawl,raffel2020exploring} and are able to tackle downstream tasks without explicit training~\citep{10.1145/3544548.3580895,10.1145/3580305.3599572,vernikos2023small}.

Despite their accuracy benefits and enabling unprecedented use-cases, such models have been pushing the computational boundaries of cloud systems, both in terms of training~\citep{dao2023flashattention} and deployment~\citep{kwon2023efficient}. As a result, specialized systems and hardware have been developed for the cloud to accelerate such demanding workloads.
Typically deployed in large data centers, this poses questions both in terms of the sustainability of such deployments~\citep{wu2022sustainable,patterson2022carbon,patterson2021carbon,zeus_2023}, as well as the privacy and custody of user data~\cite{10.1145/3524842.3528440}.
We recognize that it is not always necessary to deploy a highly over-provisioned network to solve the task at hand~\citep{eldan2023tinystories}.
Given that model performance, even for smaller models, does not saturate quickly, i.e.,~more data gives performance gains~\citep{tinyllama}, and the need for user privacy~\citep{xiao2023offsite}, we focus our attention to the study of deploying LLMs at the edge~\citep{laskaridis2022future}, with particular emphasis on the mobile execution of chat assistants.

To this end, we have created our own infrastructure, named \tool\footnote{\camready{Our code can be found at: \texttt{\href{https://github.com/brave-experiments/MELT-public}{github.com/brave-experiments/MELT-public}}.}}, designed to interact, trace and benchmark LLMs across ML frameworks, devices, and ecosystems. With our tool, we automate the interaction with instruction fine-tuned models and capture events and metrics of interest at a granular level, both in terms of performance as well as energy. While in this paper \tool is deployed for LLMs inference, our infrastructure is general enough to support the tracing of any workload, without user intervention. To the best of our knowledge, our tool is the first to support granular on-device energy measurements across device targets (i.e.,~Android, iOS, Linux) with realistic interactions.

\noindent\textbf{Research Questions.}
Given the constant developments in mobile and embedded System-On-Chips (SoCs) and the meteoric rise of LLMs, we aspire to measure the deployability of Large Language Models at the consumer edge and identify the bottlenecks that prevent the broad deployment of such workloads.
Specifically, the research questions we aim to answer with this study are the following:

\begin{enumerate}[leftmargin=*,topsep=0pt]
    \item Is it feasible to deploy LLMs locally on device in a private yet efficient manner?
    \item What is the state of inference execution across a heterogeneous ecosystem of consumer edge and mobile devices in terms of performance and energy demands? 
    \item What are the current limiting factors and bottlenecks for deploying LLMs on device?
    \item What is the impact of quantization to the performance and accuracy of the network?
    \item Given the abundance and heterogeneity of smart devices, can such workloads be realistically offloaded to ambient devices at the consumer edge?
\end{enumerate}

\noindent
Concretely, our paper makes the following contributions:

\begin{itemize}[leftmargin=*,topsep=0pt]
    \item We gather the most popular open-source LLMs and benchmark them across mid and high-tier mobile and edge devices of different manufacturers, including iOS and Android-based phones as well as Nvidia Jetson edge devices. 
    Our goal is to explore the deployability of broadly available LLMs on broadly available consumer hardware.
    \item To this end, we have developed the first mobile LLM evaluation suite, called \tool, responsible for downloading, quantizing, deploying and measuring the performance and energy of an LLM across heterogeneous targets.
    \item Through \tool, we trace specific events during inference and pinpoint their computational and energy impact. We also evaluate the continuous runtime of LLMs and their impact on battery life and user's Quality of Experience.
    \item We further quantify the impact of quantization on the accuracy of models, over different datasets and tasks.
    \item Last, we pinpoint bottlenecks in deployment and explore alternative avenues for edge deployment.
\end{itemize}

\vspace{-0.2cm}
\section{Background \& Motivation}
\label{sec:background}

\subsection{Transformer Preliminaries}

Transformers~\cite{vaswani2017attention} were introduced back in 2017 as an alternative architecture for NLP tasks, providing better performance and scalability than their recurrent counterparts and fewer inductive biases than convolutional networks. Since then, they have been expanded to more tasks, including vision~\cite{dosovitskiy2021an} and multi-modal use-cases~\cite{radford2021learning}. In this paper, we are focusing our attention on large-scale language \mbox{transformers.} %

The original transformer comprises an \textit{encoder-decoder} architecture, where the \textit{encoder} digests tokens from the input sequence, whereas the \textit{decoder} digests tokens from the output in an \textit{autoregressive} manner. Each part of the architecture consists of multiple attention blocks. There are also encoder and decoder-only model variants, which include the respective part of the architecture. Tokens are (sub-)word representations, generated by a \textit{tokenizer} model, embedded into as subspace (e.g.,~WordPiece~\cite{devlin-etal-2019-bert} or BytePair~\cite{radford2019language} \mbox{encoding}). 

The main contribution of transformers has undoubtedly been the attention mechanism, which captures the relationship between tokens in a sequence from a single source (self-attention) or multiple sources (multi-head attention). Attention is calculated as $A(Q,K,V) = \text{softmax}(\frac{QK^T}{\sqrt{d_k}})V$,
where $Q$, $K$, $V$ represent the query, key, and value matrices, respectively, and $d_k$ the dimensionality of the key matrix. The inner product boosts closer query-key vectors (relevance), softmax normalizes the dot-product and the multiplication with the value results in the relevant value scores being retrieved.
The quadratic complexity of attention, with respect to the sequence length (i.e.,~the prompt or intermediate tokens), is one of the main bottlenecks of deployment, which has given way to alternatives such as sparse~\cite{beltagy2020longformer} or approximate~\cite{wang2020linformer,ainslie-etal-2023-gqa} attention mechanisms, as well as attention-free variants as of lately~\cite{gu2023mamba}. \textit{Context size} refers to the maximal window of tokens a transformer block can pay attention to, whereas the \textit{maximum generated length} refers to the maximum number of tokens generated as output. Generation ends when an \texttt{<EOS>} (end-of-sequence) token is generated. 
The auto-regressive nature of decoding means that given an input sequence $X=\{x_1, x_2, \dots, x_t\}$, the model generates $x_{t+1}$, which is fed to the next generation step. Key-Value cache~\cite{pope2023efficiently} optimizes this by storing intermediary attention states.

\vspace{-0.2cm}
\subsection{Large Language Models}

What has made Transformers an instant success has been their applicability to various modalities and their scalability to very large parameter sizes without saturating accuracy~\cite{touvron2023llama2}. This phenomenon has given birth to Foundation Models (FMs), \textit{pretrained} on huge corpora of data, i.e.,~text in our case, and act as a great tool for modeling language and a starting point for fine-tuning on downstream tasks. The task of pretraining usually comprises masked or next-word prediction (self-supervised), whereas downstream tasks can include anything from translation to summarization. \textit{Instruction fine-tuning~\cite{ouyang2022training}} refers to a specific form of fine-tuning where the model is trained on pairs of input-output instructions.
Last, \textit{alignment} is usually the final step of model tuning, typically through reinforcement learning from human~\cite{ouyang2022training} or automated~\cite{bai2022constitutional} feedback, to promote a certain style or content or response that ``aligns'' with values of the creator (e.g.,~safety). Training cost generally scales down as we move from pretraining downstream, as do data ingestion needs~\cite{zhou2024lima}.

\vspace{-0.2cm}
\subsection{Current State and Motivating Factors}

\noindent
\textbf{Centralization and privacy.} Training a large-scale LLM is a costly effort, and many models are only offered as black-box solutions to users, such as ChatGPT and GPT-4 by OpenAI, Claude by Anthropic or Gemini by Google. These are offered as-a-service, which means that user prompts are transmitted to the provider, thereby compromising user-privacy. At the same time, users lack control over whether their data get incorporated in the training set of models without their explicit consent~\cite{10.1145/3524842.3528440}, making them amenable to various attacks~\cite{nasr2023scalable}. Additionally, these tools remain accessible and operational only under an active internet connection.

\noindent
\textbf{LLMs democratization.} Nevertheless, more and more models offer openly
their weights, including models from Meta~\cite{touvron2023llama,touvron2023llama2}, Mistral AI~\cite{jiang2023mistral}, Google~\cite{gemma} and Microsoft~\cite{phi2}. This creates an excellent opportunity for users to deploy their models locally and even personalize them to their preferences, without data ever leaving their device premises. However, such models remain significantly smaller in scale and still require considerable resources to deploy. Toward this end, new frameworks are emerging for enabling local execution of LLMs across different targets~\cite{llama.cpp,mlc-llm,mlx2023,mediapipe,MNN-LLM,tinygrad}. In this effort, quantization~\cite{shen2020q,frantar2022gptq,lin2023awq} is one of the most prominent out-of-the-box solutions for reducing their footprint. Yet another enabler towards this democratization is the broad availability of capable SoCs at cost. Indicatively, from our measurements, a recent M2-based Mac Studio can run Llama-2~\cite{touvron2023llama2} 7B model (4-bit quantized) at a sustained 46.8 tokens/sec.

\noindent
\textbf{Sustainability.} Last but not least, the issue of sustainability becomes ever more pronounced~\cite{wu2022sustainable,patterson2021carbon,patterson2022carbon,zeus_2023}, since the training and deployment of large models requires a significant amount of energy, be it inside or outside the premises of the data center. As a result, the cost is not only monetary, but also energy consumption bound.

For all the reasons above, we feel it is more critical than ever before to quantify the cost or running LLMs on mobile and edge devices, the current bottlenecks and the sustainability of this deployment model. This way we aim to fuel future research avenues for optimizing local model deployment and further democratizing their adoption.

\vspace{-0.2cm}
\section{\tool Infrastructure}
\label{sec:melt}
\begin{figure*}[t]
    \vspace{-0.6cm}
    \centering
    \begin{subfigure}[t]{0.5\textwidth}
        \centering
        \includegraphics[width=0.95\linewidth,trim={14.8cm 8cm 14.8cm 6cm},clip]{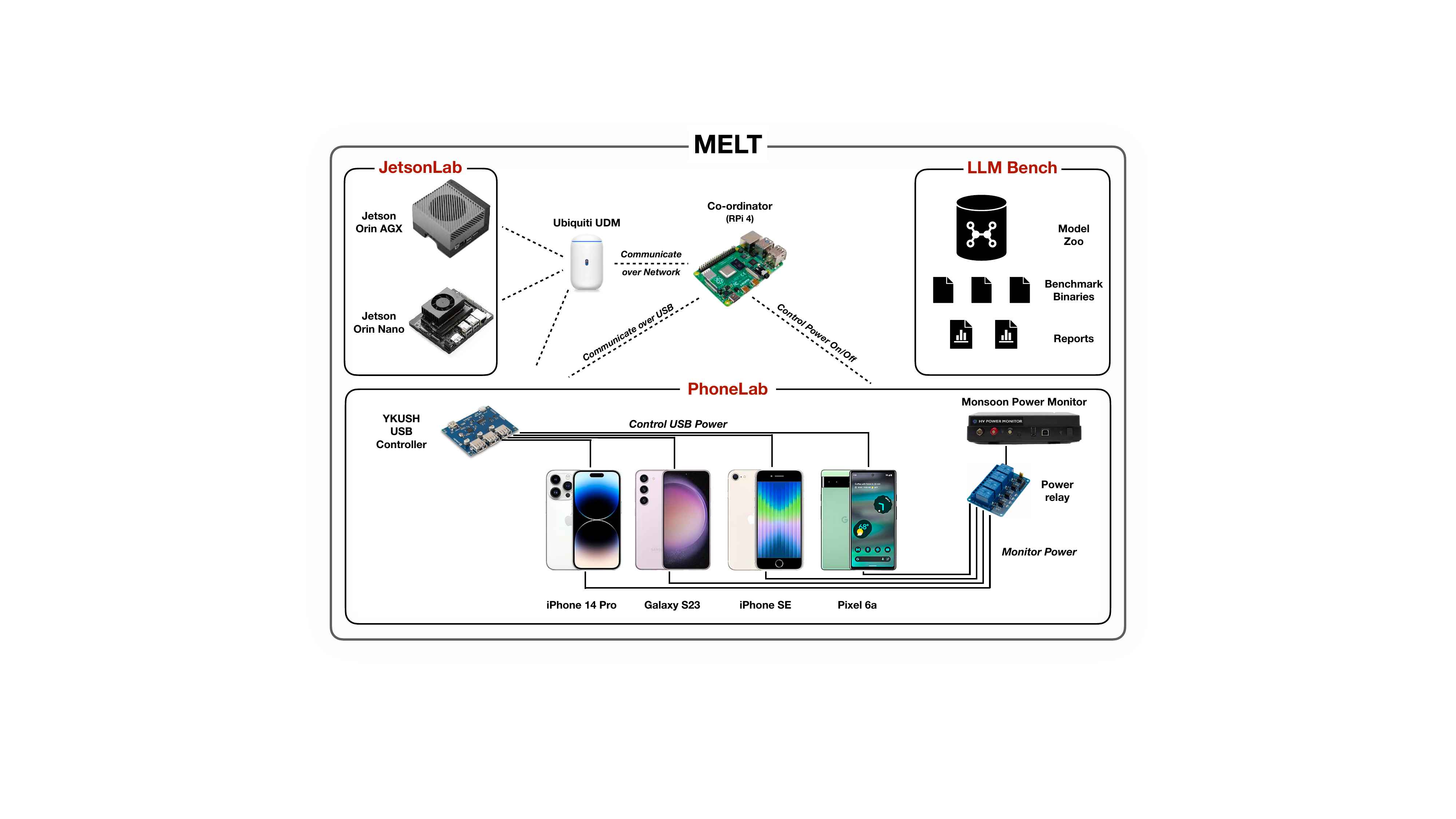}
        \vspace{-0.2cm}
        \caption{Architecture of MELT device farm}
        \label{fig:melt-arch}
    \end{subfigure} \hspace{0.1cm}
    \begin{subfigure}[t]{0.42\textwidth}
        \centering
        \includegraphics[width=\linewidth,trim={14.8cm 9.5cm 14.8cm 5cm},clip]{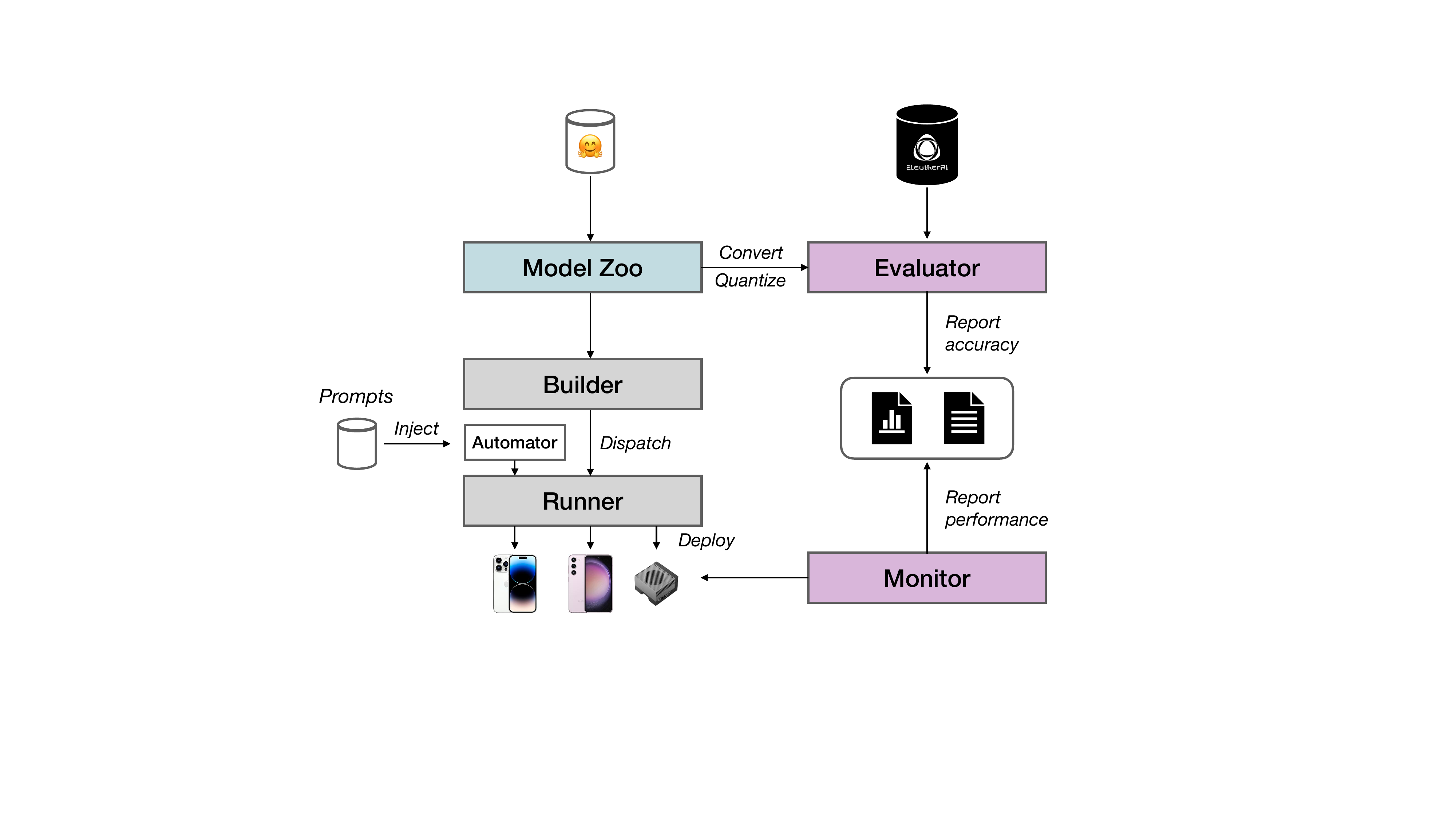}
        \vspace{-0.5cm}
        \caption{MELT Workflow}
        \label{fig:melt-workflow}
    \end{subfigure}
    \vspace{-0.4cm}
    \caption{Architecture and workflow of \tool}
    \vspace{-0.4cm}
\end{figure*}

In order to benchmark the runtime of LLMs on edge and mobile devices, we have engineered our own device farm, which comprises a combination of hardware and software components working in tandem to automate and measure robustly the on-device behavior of the targeted use-case. 
Our infrastructure adopts a client-server architecture, with the \emph{coordinating} process running on a Raspberry Pi 4 (RPi).
This is responsible for \textit{i)}~organizing the execution of the benchmarking suite, \textit{ii)}~scheduling and dispatching jobs to devices, \textit{iii)}~controlling downstream interaction with the application, \textit{iv)}~monitoring their runtime, temperature and energy consumption along with the \textit{v)}~tracing the events of interest in the downstream task.

The \emph{co-ordinator} communicates with two sets of devices, namely \emph{PhoneLab} (Sec.~\ref{sec:phonelab}) which consists of mobile devices and \emph{JetsonLab} (Sec.~\ref{sec:jetsonlab}), which includes Nvidia Jetson boards, as the name suggests.
We support the monitoring and interaction with Graphical User Interface (GUI) and Command Line Interface (CLI) applications, so that performance can be measured in realistic settings as well as in silo.

\begin{table}[t]
    \centering
    \caption{Device Farm of \tool}
    \vspace{-0.4cm}
    \begin{adjustbox}{width=1.0\linewidth}
        \begin{tabular}{l p{3.4cm} c c p{1.9cm} c c }
        \toprule
        \textbf{Device Model} & \textbf{SoC} & \textbf{Mem.}  & \textbf{Battery} & \textbf{OS version} & \textbf{Year} & \textbf{Tier} \\
        \midrule
        \multicolumn{7}{l}{\textbf{Co-ordinator \& Builder}} \\
        {Raspberry Pi 4} & Broadcom BCM2711 & 8GB & - & RPi OS 11.9 & 2019 & - \\
        {Mac Studio} & M2 Max & 32GB & - & macOS 14.1.2 & 2023 & - \\
        \midrule
        \multicolumn{7}{l}{\textbf{Mobile}} \\
        {Galaxy S23} & Snapdragon 8 Gen 2 & 8GB & 3785 mAh & Android 14 & 2023 & High \\
        {Pixel 6a} & Tensor Core & 8GB & 4410 mAh & Android 13 & 2023 & Mid \\
        {iPhone 14 Pro} & A16 Bionic & 6GB & 3200 mAh & iOS 17.3.1 & 2022 & High \\
        {iPhone SE} & A15 Bionic & 4GB & 1821 mAh & iOS 17.3.1 & 2022 & Mid \\
        \midrule
        \multicolumn{7}{l}{\textbf{Edge}} \\
        {Jetson Orin AGX} & NVIDIA Carmel + \mbox{Ampere GPU} & 64GB & - & Ubuntu 20.04 (L4T 35.2.1) & 2022 & High \\
        {Jetson Orin Nano} & 8-core Arm Cortex-A78AE + Ampere GPU & 8GB & - & Ubuntu 20.04 (L4T 35.4.1) & 2022 & Mid \\
        \bottomrule
        \end{tabular}
        \label{tab:device-farm}
    \end{adjustbox}
    \vspace{-0.4cm}
\end{table}

\vspace{-0.2cm}
\subsection{PhoneLab}
\label{sec:phonelab}

We have incorporated four smartphones into our device farm, spanning across different resource tiers (mid and high tier) and platforms (Android and iOS), as detailed in Tab.~\ref{tab:device-farm}. These mobile devices are interfaced with a Monsoon high-voltage power monitor (model AAA10F)~\cite{monsoon}. To facilitate accurate power measurements, we employ a battery bypass process that requires disassembling each device to remove its battery, extracting the internal battery controller and expose the power terminals through cables. This setup ensures precise monitoring of the devices' power consumption directly from their power terminals~\cite{varvello2022batterylab} at a maximum frequency of 5KHz through Monsoon.
In order to support the powering of multiple devices, we have a \emph{programmable relay} that communicates over general-purpose input/output (GPIO) pins of Raspberry Pi and can selectively power on and off the devices, one at a time. The host machine initially communicates with the mobile devices via USB, connected over a \textit{YKUSH Switchable Hub}~\citep{ykush}. Its purpose is to selectively disable the power lanes of the USB connection, so as not to measure USB charging draw. For monitoring the thermal behavior of the devices, we have a Flir One Edge wireless \emph{thermal camera} positioned at 0.5-1.0m from the device whose temperature we want to measure. 
To minimize the influence from extraneous factors we disabled the automated OS and App updates, turned off the adaptive brightness/charging/battery features, enabled the dark mode and standardized the brightness level to 25\% across devices.
We call this part of the infrastructure \emph{PhoneLab} (see Fig.~\ref{fig:melt-arch}).

Communication to Android devices is accomplished via the Android Debug Bridge (ADB). This enables us to interact (over tap or typing events) over CLI commands with the device and application, without the need for explicit human intervention during the experiment.
ADB connection is established over Wi-Fi 6 (5GHz channel) for automation, because data and power lines cannot be independently controlled over the USB channel. 
Interfacing with iOS is more intricate, as there is no automated toolchain for controlling the device. To achieve this, we have built a Python-based service which maps commands like touch, swipe, and text input to a virtual Human Interface Device (HID), simulating a Bluetooth keyboard and mouse that controls the device. In both cases, the baseline power draw of Bluetooth and Wi-Fi events is subtracted from the energy traces.
For the compilation and deployment of apps, we have a Mac Studio in the same network as the rest of PhoneLab, with remote access to the devices.
Packages are installed through ADB and \texttt{ideviceinstaller}~\cite{ideviceinstaller} for Android and iOS, respectively.

\vspace{-0.2cm}
\subsection{JetsonLab}
\label{sec:jetsonlab}

At the same time, the \emph{co-ordinator} is connected over Ethernet to the same network as our Jetson boards with SSH access to them. We are able to take power and temperature metrics through SysFS probes available on the devices, at a frequency of approximately 100Hz\footnote{This granularity was explicitly tuned to capture events of interest, without interfering with the measurement itself due to I/O thrashing.}. This way, not only can we calculate the power and thermal behavior of each device, but we are also able to calculate the power draw from specific components of the board (e.g., CPU, GPU, SoC, DRAM, etc.). Last, Jetson devices support a range of predefined power modes, which we control over the \texttt{nvpmode}. For all experiments, we used the fan speed in its maximum setting. We call this part of the infrastructure \emph{JetsonLab} (see Fig.~\ref{fig:melt-arch}).

Compilation of packages and models happens directly on Jetson devices over Docker images\footnote{Based on images from \url{https://github.com/dusty-nv/jetson-containers/}.}. Automation is handled over SSH commands from RPi and results are collected immediately after execution. Both Jetsons have their Operating System (OS) installed on a high speed UHS-I SD card
and have dedicated M2 SSDs for the rest of the filesystem, where models and executables reside.

\SetArgSty{textsmall}
\SetAlFnt{\scriptsize}
\SetAlCapFnt{\scriptsize}
\SetAlCapNameFnt{\scriptsize}
\setlength{\textfloatsep}{0pt}
\begin{algorithm2e}[h]
    \SetAlgoLined
	\LinesNumbered
	\DontPrintSemicolon
        \nonl \textit{Pseudocode for MELT experiments. Functionality of undefined methods in comment. Prefixed methods run on the device in prefix (e.g., Monsoon, device).}\;
	\KwIn{PhoneLab, JetsonLab, Monsoon, GPIO, YKUSH, device, $Q^\text{device}_\text{experiments}$, iterations, samplingFrequency, betweenExpSleep}
        PowerOn(device) \label{alg-line:poweron}\;
        \If{device.platform == "ios"}{
            ConnectBT(device) \textcolor{blue}{\# connect as HID device via Bluetooth} \;
            UnlockScreen(device) \textcolor{blue}{\# unlock screen with passcode over HID} \;
        }
        SyncClocks(device) \textcolor{blue}{\# sync host and guest clocks} \label{alg-line:clock-sync}\;
        apiAddress = StartRESTServer() \textcolor{blue}{\# start REST service on host} \;
	\For( \textcolor{blue}{\# iterate over experiments in the queue}){exp in $Q^\text{device}_\text{experiments}$}{
            Push([exp.model, exp.conversations], device) \textcolor{blue}{\# push dependencies} \label{alg-line:job-deps} \;
            Apply(exp.conf, exp.model, device) \textcolor{blue}{\# edit model conf and execution parameters on device}\;
            \For{it=0; it<iterations;++it}{
                StartMonitoring(Monsoon, device)\;
                RunExperiment(exp, device) \label{alg-line:run-experiment}\;
                StopMonitoring(Monsoon, device) \textcolor{blue}{\# disable monitoring}\;
                CollectMeasurements(exp, device) \textcolor{blue}{\# get results from FS} \label{alg-line:collect-results}\;
                sleep(betweenExpSleep) \textcolor{blue}{\# sleep between runs} \label{alg-line:sleep-between-runs}\;
            }
	}

    \SetKwProg{Fn}{def}{:}{end}
    \Fn{PowerOn(GPIO, YKUSH, device)}{
        \If {device in PhoneLab.devices}{
            GPIO.EnableRail(device.rails) \textcolor{blue}{\# enable rail through GPIO}\;
            YKUSH.PowerOn(device) \textcolor{blue}{\# enable YKUSH USB of device}\;
            Monsoon.SetVoutCurr(device) \textcolor{blue}{\# configure Monsoon power out}\;
            Wait(device) \textcolor{blue}{\# wait until device is responsive}\;
        }
    }
    \Fn{StartMonitoring(Monsoon, YKUSH, device)}{
        \If {device in PhoneLab.devices}{
            YKUSH.DisableUSB()\;
            Monsoon.MeasurementMode("on", samplingFrequency)\;
        }
        \ElseIf {device in JetsonLab.devices} {
            Jetsonlab.ScheduleEvents(samplingFrequency)\;
            Jetsonlab.Monitor("on") \textcolor{blue}{\# poll SysFS}\;
        }
    }
    \Fn{RunExperiment(exp, device, apiAddress)}{
        \textcolor{blue}{\# open app w/ ADB, Bluetooth HID or SSH} \;
        app = device.OpenApp(exp.backend)\;
        Automate(app, model, device) \textcolor{blue}{\# automate interaction with app} \;
        http.post("start", apiAddress) \textcolor{blue}{\# notify through REST service} \;
        \For{conversation in exp.conversations}{
            \For{prompt in conversation} {
               report = device.Trace(model(prompt)) \textcolor{blue}{\# run inference} \; 
               device.Write(report, exp.conf.outputPath) \textcolor{blue}{\# results to FS} \;
            }
        }
        http.post("stop", apiAddress) \textcolor{blue}{\# notify through REST service} \;
    }
\caption{\footnotesize \mbox{\textbf{\tool} (Experiment Process)}}
\label{alg:run_experiment_algo}
\end{algorithm2e}

\vspace{0.2cm}
\section{Methodology}
\label{sec:methodology}

For the purpose of measuring LLMs performance on device, we created \tool as a benchmarking framework, which is responsible for i)~the download and conversion/quantization of models, ii)~the compilation of the respective benchmarking suite backend, iii)~the deployment, automation and runtime of the LLM on the respective device, iv)~the fine-grained monitoring of resource and energy consumption of the execution and v)~the reporting of the results. The workflow of \tool is depicted in Fig.~\ref{fig:melt-workflow}.

\subsection{Model Zoo and Evaluation}

\noindent\textbf{Model Zoo.}
As a first step, we collect the models we would like to benchmark on device from their respective sources and convert them, based on the backends available, to the respective format (e.g., GGUF - formerly known as GGML - for llama.cpp; MLC/TVM compiled files and libraries for MLC-LLM). The benchmarked models are shown on Tab.~\ref{tab:supported-models}. Moreover, given the sheer size of the model weights, more often than not, it is necessary to quantize the models to lower precision so that their memory footprint is reduced, and the traffic between on-chip and DRAM memory is smaller.
To this end, \tool's converter is able to resolve and download models from git or huggingface and convert their weights to the respective format.
This format varies both in terms of the ML framework, as well as the hardware executing the network. The supported formats and quantization methods are depicted in Tab.~\ref{tab:llm-frameworks}. The original models were downloaded directly from HuggingFace Hub and the converted models reside in \tool's \textit{Model Zoo}, which is a repository of converted models available to be benchmarked.

\noindent\textbf{Model Evaluator.}
The next step is to evaluate the accuracy degradation of the model due to quantization. To accomplish this, we use \tool's \textit{Model Evaluator} component, which is responsible for evaluating the model\footnote{We evaluate the non-finetuned variants of the models, as a typical proxy of the accuracy degradation of downstream models.} on a given dataset and reporting its accuracy.
We leveraged the LM-Evaluation Harness~\cite{eval-harness} and integrated a custom inference server to serve our converted models from each of the supported backends. This offers a convenient abstraction layer between the frameworks and the evaluation harness. 
Because of the lack of native support from the frameworks, we had to implement the extraction of token log probabilities to assess the accuracy per downstream dataset\footnote{Because of issues with evaluating quantized models on MLC-LLM, we evaluate AWQ~\cite{lin2023awq} quantized models with \texttt{autoawq} package as a proxy.}. 
The currently supported datasets are depicted in Table~\ref{tab:evaluation_datasets} and the results of the evaluation are presented in Sec.~\ref{sec:quantization_acc}.

\begin{table}[t]
\centering
\caption{Supported pretrained models}
\vspace{-0.4cm}
\label{tab:supported-models}
\begin{adjustbox}{width=\linewidth}
\begin{tabular}{l l l l}
\toprule
\textbf{Model Type} & \textbf{Size} & \textbf{Type} & \textbf{HuggingFace Repository} \\
\midrule
\multirow{1}{*}{\textbf{TinyLlama}}~\citep{tinyllama} & 1.1B & Decoder & \href{https://huggingface.co/TinyLlama/TinyLlama-1.1B-Chat-v0.5}{\textit{TinyLlama/TinyLlama-1.1B-Chat-v0.5}} \\
\multirow{1}{*}{\textbf{Zephyr-3B}} & 3B & Decoder &  \href{https://huggingface.co/stabilityai/stablelm-zephyr-3b}{\textit{stabilityai/stablelm-zephyr-3b}} \\
\multirow{1}{*}{\textbf{MistralAI-7B}} & 7B & Decoder &  \href{https://huggingface.co/mistralai/Mistral-7B-Instruct-v0.1}{\textit{mistralai/Mistral-7B-Instruct-v0.1}} \\
\multirow{2}{*}{\textbf{Gemma}~\citep{gemma}} & 2B & \multirow{2}{*}{Decoder} & \href{https://huggingface.co/google/gemma-2b-it}{\textit{google/gemma-2b-it}} \\
& 7B &  & \href{https://huggingface.co/google/gemma-2b-it}{\textit{google/gemma-7b-it}} \\
\multirow{2}{*}{\textbf{Llama-2}~\citep{touvron2023llama2}} & 7B & \multirow{2}{*}{Decoder} & \href{https://huggingface.co/meta-llama/Llama-2-7b-chat-hf}{\textit{meta-llama/Llama-2-7b-chat-hf}} \\
& 13B &  & \href{https://huggingface.co/meta-llama/Llama-2-13b-chat-hf}{\textit{meta-llama/Llama-2-13b-chat-hf}} \\
\bottomrule
\end{tabular}
\end{adjustbox}
\end{table}

\subsection{Automated On-Device Benchmarking}

\noindent\textbf{Benchmark Workflow.}
During the execution of the respective model, we have instrumented the binaries of each framework so that we can report fine-grained timings of chat and model operations. This instrumentation includes timing of granular chat and DNN graph operations as well as calculation of performance metrics. Chat events include operations such as \textit{prefill}, \textit{encoding} or \textit{decoding}, whereas graph operations refer to the LLM layers and kernel operations, which vary per framework because of optimizations happening during model conversion (e.g.,~operator fusion~\cite{chen2018tvm}). Due to the overhead of tracing very granular events (i.e.,~single operations), we only enable the respective flag in specific experiments (Sec.~\ref{sec:microbenchmarks}).

\noindent\textbf{Builder.}
In order to evaluate the performance across devices, we have used two frameworks that have constituted so far the benchmarks for executing LLMs on device, namely MLC-LLM~\cite{mlc-llm, chen2018tvm} and llama.cpp~\cite{llama.cpp} (detailed in Tab.~\ref{tab:llm-frameworks}). While there are increasingly more such frameworks~\cite{MNN-LLM,mlx2023,mediapipe,tinygrad,han_tinychatengine}, we selected the ones with the highest popularity (measured by their stars on GitHub) and widest model and platform support. We have made \tool extensible so that new frameworks can be integrated with minimal effort.

We have automated the build of the framework backends and applications for each platform  (e.g.,~Android, iOS, Linux (CUDA)), along with the conversion binaries for the respective models. We used an M2-powered Mac Studio on the local network to build and package dependencies for mobile targets, especially since Xcode was required to sign app releases on iOS. Specifically, the Android apps were built with Android SDK v.35.0.0 and NDK v.26.1, whereas for iOS we used Xcode 15.2. Installation of packages (\texttt{.apk} and \texttt{.ipa}) was done by the co-ordinator. For the case of \emph{JetsonLab}, the frameworks and models were compiled on device with CUDA 12.2.

\noindent\textbf{Automator.}
In order to measure the performance of the respective model on device, we automate the interaction with the chat application. To accomplish this, we use a set of precanned prompts, sampled from the OAAST chat dataset~\cite{kopf2023openassistant},
and interact in a multi-turn manner with the LLM. 
More information on the distribution of these prompts in Sec.~\ref{sec:dataset_qualitative_analysis}.

For mobile execution, we have used custom native applications\footnote{All applications have graphical user interface except for llama.cpp on Android, for which we used the ADB CLI interface~\cite{almeida2021smart}.} that automatically read prompts from a given file and replay the discussion with the model at hand.
For edge execution and Android llama.cpp, we leverage the command-line interface to converse with the LLM and automate the interaction with \texttt{expect} scripts. These are \texttt{TCL}-based scripts that operate based on the text output of a binary. In the future, we would also like to evaluate guardrail chat mechanisms~\cite{rebedea-etal-2023-nemo} and how the impact runtime characteristics.

For \emph{JetsonLab}, transferring the dependencies and executing the job is accomplished over SSH commands. For \emph{PhoneLab}, the process is more involved.  For Android devices, communication and execution of jobs is mostly handled over {ADB}. We use the {ADB} as the controller for transferring files, installing and launching the application as well as automating the interaction with the app (i.e.,~launching a fragment or tapping on screen elements). For iOS devices, we emulate an HID Bluetooth device with the RPi that acts as a combo mouse/keyboard device. This way, we carefully script the series of actions that need to be taken so that we launch and execute a job on that device. 
At the end of the experiment, the co-ordinator (RPi) is automatically notified when the evaluation task is complete through a REST request. The reason behind this is for the co-ordinator to know when an experiment has finished to stop energy measurements, persist logs and continue with the next job. At the same time, we collect the generated responses and the metrics of interest. 

\begin{table}[t]
        \caption{Frameworks and platforms supported by \tool.}
    \vspace{-0.4cm}
    \begin{adjustbox}{width=\linewidth}
        \begin{tabular}{l l l p{3.8cm} p{4.5cm}}
        \toprule
        \textbf{Framework} & \textbf{Backend} & \textbf{Version} & \textbf{Supported Platforms} & \textbf{Quantization} \\
        \midrule
        \textbf{MLC-LLM}~\cite{mlc-llm} & TVM~\cite{chen2018tvm} & \texttt{96a68e}$^\dagger$ & Android (GPU), iOS (Metal), Linux (CUDA) & Group Quantization~\cite{shen2020q}, GPTQ~\cite{frantar2022gptq}, FasterTransformer Row-wise Quantization \\
        \textbf{llama.cpp}~\cite{llama.cpp} & llama.cpp~\cite{llama.cpp} & \texttt{b22022}$^\ddagger$ & Android (CPU, GPU), Linux (CUDA) & \multirow{3}{*}{k-quants~\cite{k-quants}}\\
        \textbf{LLMFarm}~\cite{LLMFarm} & llama.cpp~\cite{llama.cpp} & \texttt{7226a8}$^*$ & iOS (Metal) & \\
        \bottomrule
        \multicolumn{5}{l}{$^\dagger$ We used version \texttt{784530} for supporting Gemma models and Llama-2-7B on Android.} \\
        \multicolumn{5}{l}{$^\ddagger$ We used version \texttt{d5ab29} for supporting Gemma models.} {\camready{$^*$ We used version \texttt{46bdb4} for supporting Gemma models.}}
        \end{tabular}
        \label{tab:llm-frameworks}
    \end{adjustbox}
\end{table}

\noindent\textbf{Runner.}
The runner is tasked with deploying the built application or binary, along with the associated converted models to the respective device, running the automated interaction and gathering the reported results and logs. The experiment runtime is documented in more detail in Algorithm~\ref{alg:run_experiment_algo}.
When an experiment is run, the \emph{co-ordinator} is responsible for powering the device if in PhoneLab (L.\ref{alg-line:poweron}), connecting to it (over SSH or USB), synchronizing the clocks (L.\ref{alg-line:clock-sync}), deploying the job dependencies (model, application, inputs) (L.\ref{alg-line:job-deps}), executing the task (L.\ref{alg-line:run-experiment}) and gathering the outputs to return (L.\ref{alg-line:collect-results}). This happens over multiple iterations, with configurable waiting times between experiments (L.\ref{alg-line:sleep-between-runs}). 

\begin{figure*}[t!]
    \vspace{-0.4cm}
    \begin{subfigure}[t]{0.54\textwidth}
        \centering
        \begin{subfigure}[t]{.4\textwidth}
        \centering
        \includegraphics[width=0.8\linewidth]{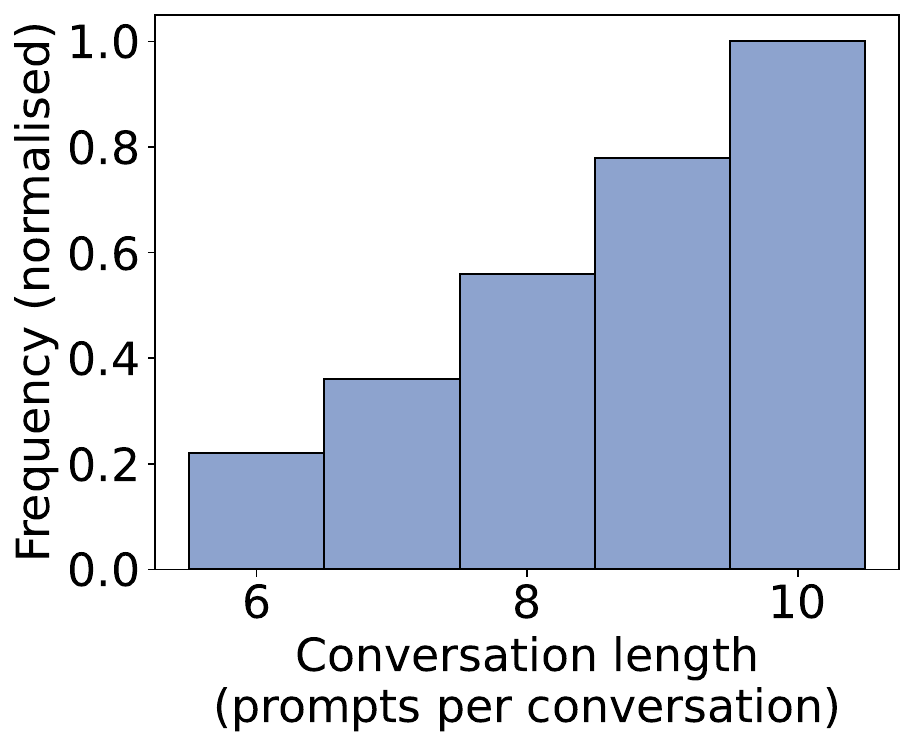}
        \end{subfigure}
        \begin{subfigure}[t]{.4\textwidth}
        \centering
        \includegraphics[width=0.8\linewidth]{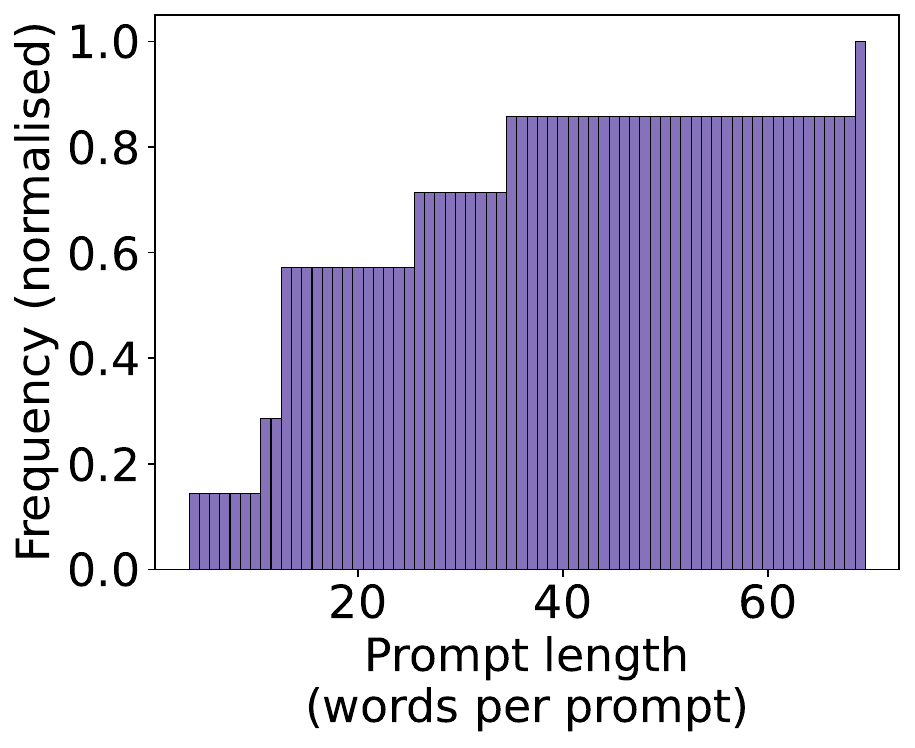}
        \end{subfigure}
        \vspace{-0.2cm}
        \caption{CDFs of conversation (\# prompts) and prompt lengths (\# words)}
        \label{fig:len-histograms}
    \end{subfigure}\hspace{0.1cm}
    \begin{subfigure}[t]{0.44\textwidth}
        \centering
        \includegraphics[width=.55\linewidth]{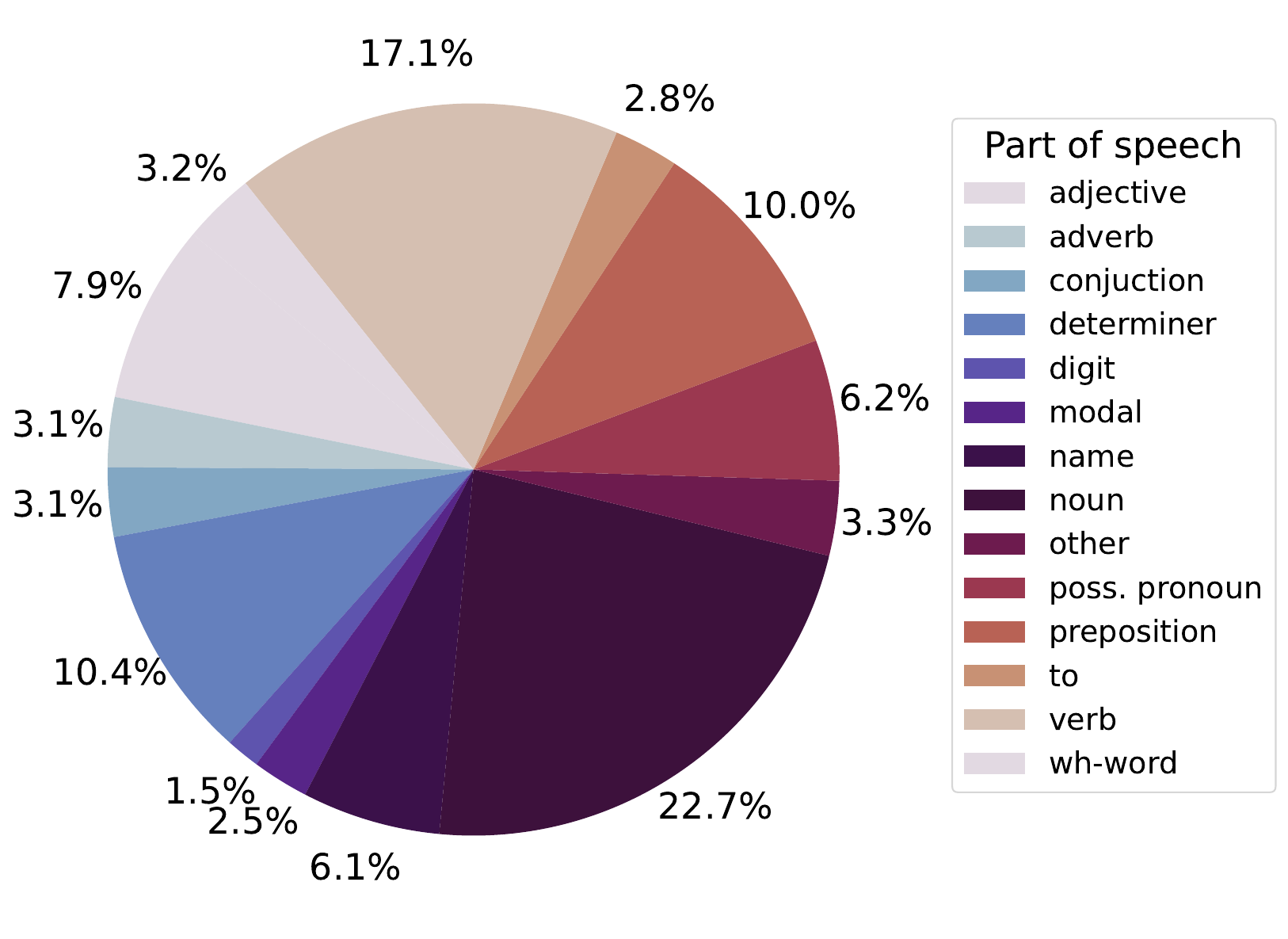}
        \vspace{-0.2cm}
        \caption{Part of speech categories distribution across prompts}
        \label{fig:pos-pie}
    \end{subfigure}
    \vspace{-0.4cm}
    \caption{Qualitative analysis of prompts used for macro-experiments to assess the behaviour of LLM-powered chats on device.}
    \vspace{-0.4cm}
    \label{fig:prompt-qual}
\end{figure*}

\noindent\textbf{Monitor.}
Our monitoring infrastructure comprises a combination of hardware and software components. We measure \textit{coarse} (end-to-end) and \textit{fine-grained} (per-operator) metrics about latency and memory from the benchmark binaries. We also traced the execution through Android, Xcode and Nvidia Visual profilers for analyzing the behavior of each runtime across different platforms. These were invoked in isolation due to their overhead.
These give us computational information about the LLM workload. At the same time, as aforementioned in Sec.~\ref{sec:melt}, our mobile devices from PhoneLab are connected to a Monsoon high-voltage power monitor (AAA10F) for energy measurements, while JetsonLab supports power monitoring through SysFS probes.
These metrics are buffered in memory and asynchronously persisted to the filesystem in a CSV timeseries file. As we have granular and synchronized timings for each operation of the LLM chat execution, we can correlate the power and thermal behavior of the device with the execution of the respective operation.

\vspace{-0.2cm}
\section{Evaluation}
\label{sec:evaluation}
In this section, we present results from running LLMs across devices and platforms with \tool.
We start by describing our experimental setup and how we have run our experiments in Sec.~\ref{sec:experimental_setup}. Next, we move to the on-device evaluation of various models, showcasing the computational, memory, energy and thermal behavior of these workloads.
Specifically, we can distinguish our measurements in two settings: \textit{i)}~\emph{macro-experiments}, where we measure how a chat assistant behaves on device, with real conversations (details in Sec.~\ref{sec:dataset_qualitative_analysis}) and variable token length output, and \textit{ii)}~\emph{micro-experiments}, where we fix the output length and disregard \texttt{<EOS>} tokens, so that we measure specific operations in a more controlled manner. The former setting aims to quantify the realistic behavior of chat assistants while the latter is destined for specific operation tracing.
Since we heavily employ quantization to deploy LLMs on device, we also quantify its impact on various language tasks in Sec.~\ref{sec:quantization_acc}. Last, understanding that the constant release of new models and the fact that current generation of mobile hardware may not be yet optimized for this type of workloads, we explore in Sec.~\ref{sec:jetson_evaluation} the performance of running LLMs when deployed on local edge devices, i.e.,~Nvidia Jetson devices, and comment on the potential of edge offloading.

\subsection{Experimental Setup}
\label{sec:experimental_setup}

For our experiments, we leverage the infrastructure and methodology described in Sec.~\ref{sec:melt} and \ref{sec:methodology}, respectively. For each device (Tab.~\ref{tab:device-farm}), we tweak the {model size}, {quantization bitwidth}, {context size}, {maximum generated length} and {token batch size} through a grid search\footnote{(context size=\{512, 1024, 2048\} $\odot$ max gen. length=\{64, 128, 256\}) $\times$ batch size=\{128, 512, 1024\}, where $\odot$ is the Hadamard and $\times$ the Cartesian product.}. GPU acceleration has been used for MLC-LLM and llama.cpp when it yielded performance benefits.
This was not the case for llama.cpp on Android, where the gains from running on GPU were minimal\footnote{Indicatively, running TinyLlama-1.1B (4-bit) on S23 resulted in {13.61$\pm$0.54} vs. {13.22$\pm$0.46} tok/sec on CPU and GPU, respectively. Others have also documented this: \url{https://github.com/ggerganov/llama.cpp/issues/5965}.}. \camready{As such, for comparative performance, we have shown CPU runtimes for llama.cpp on PhoneLab.} We based our infrastructure on the versions of frameworks shown on Tab.~\ref{tab:llm-frameworks}, but with further instrumentation and automations on our side to support the scalable evaluation of performance across platforms and devices. We used the models of Tab.~\ref{tab:supported-models}, and converted/quantized them with the native tools of each backend. This was necessary as we needed to alter the generated libraries for instrumentation. Unless stated otherwise, all experiments were repeated three times and we report mean and standard deviation of the runs.

\subsection{Macro Experiments}
\label{sec:macro-experiments}

\subsubsection{Dataset Qualitative Analysis}
\label{sec:dataset_qualitative_analysis}

For macro-experiments, we used a subset of prompts from the \mbox{OpenAssistant/oasst1} dataset~\cite{kopf2023openassistant}. We filtered out inputs, so that the resulting dataset has prompts in English, with at least 5 turns of interaction. We used a sample of 2k data points and ended up with a dataset of 50 conversations. We present some qualitative results on Fig.~\ref{fig:prompt-qual}, where we depict the distributions of conversation lengths, prompt lengths and also part-of-speech categories across prompts. We can see from Fig.~\ref{fig:len-histograms} that the conversation length spans linearly from 6 to 10 prompts with the 80-th percentile of prompts below 36 words. Most words represent verbs, determiners and nouns, as analyzed with the \texttt{nltk} python package. We combined the long tail of tags of less than 1\% to the category ``other''. Of course, the correspondence of words to tokens depends on the tokenizer used by the respective model.

\subsubsection{On-device Runtime} 
\label{sec:on-device-runtime}
We start by quantifying the token throughput and efficiency per device and framework.

\begin{figure}[t]
    \centering
    \begin{subfigure}[t]{0.23\textwidth}
        \centering
        \includegraphics[width=\linewidth,trim={0 1cm 0 0},clip]{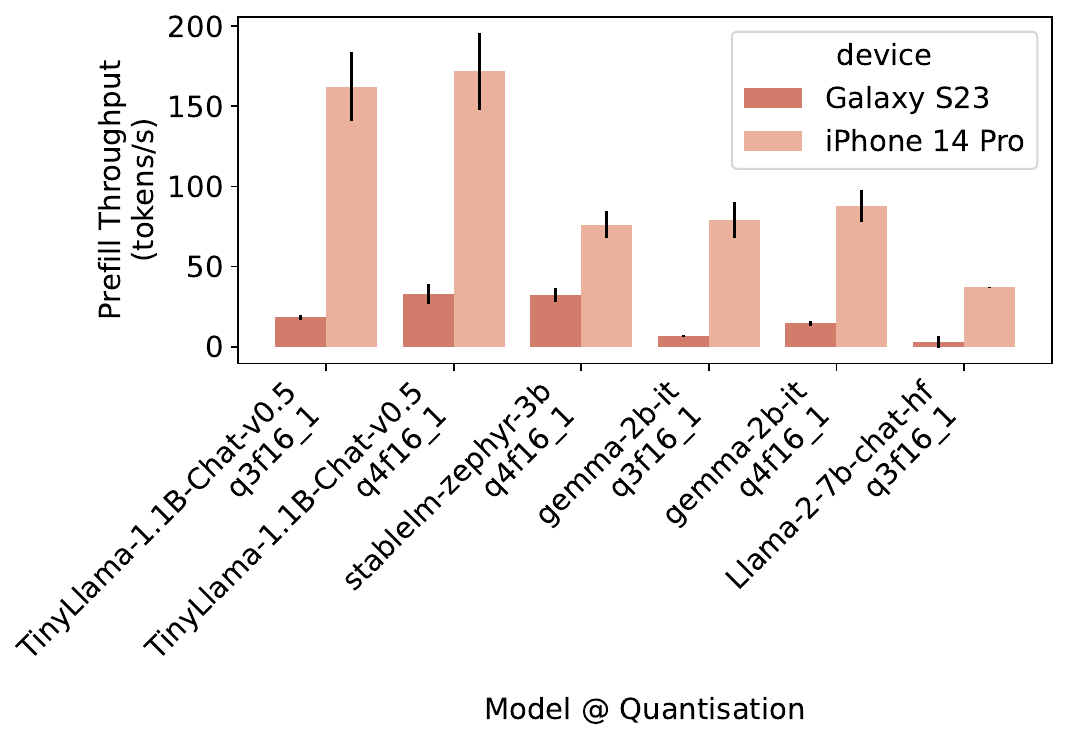}
        \caption{Prefill throughput for \mbox{MLC-LLM} on high-tier devices (GPU)}
        \label{fig:prefill_mlc_hightier}
    \end{subfigure}
    \begin{subfigure}[t]{0.23\textwidth}
        \centering
        \includegraphics[width=\linewidth,trim={0 1cm 0 0},clip]{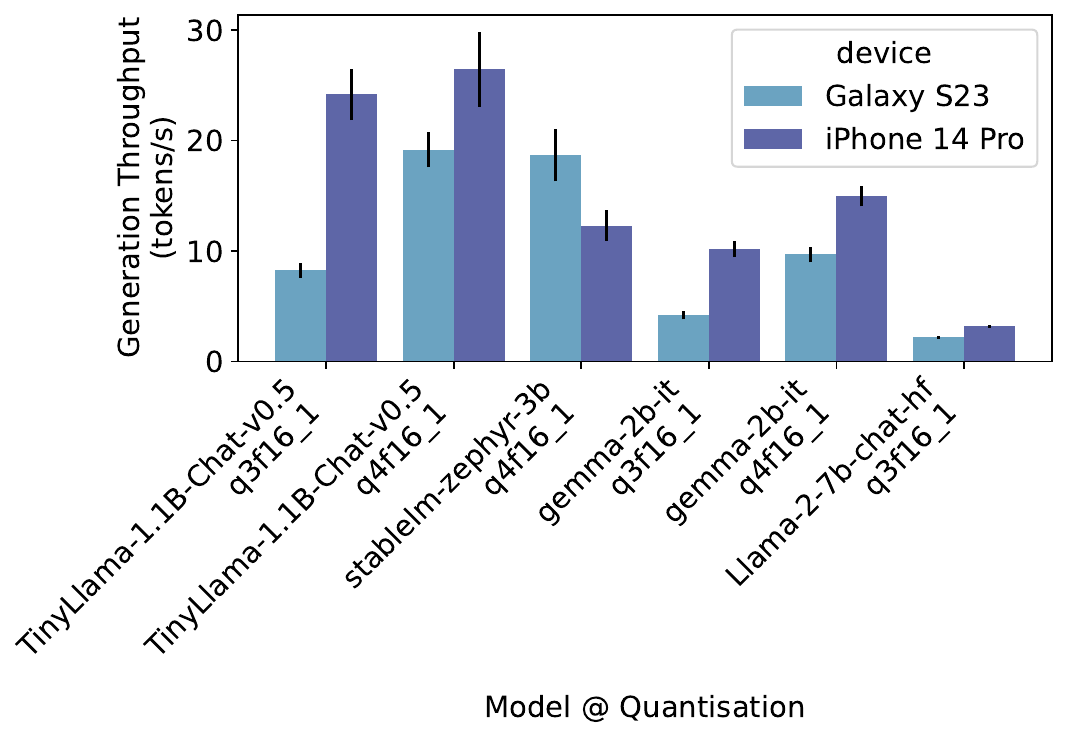}
        \caption{Generation throughput for \mbox{MLC-LLM} on high-tier devices (GPU)}
        \label{fig:gen_mlc_hightier}
    \end{subfigure}
    \begin{subfigure}[t]{0.23\textwidth}
        \centering
        \includegraphics[width=\linewidth,trim={0 1cm 0 0},clip]{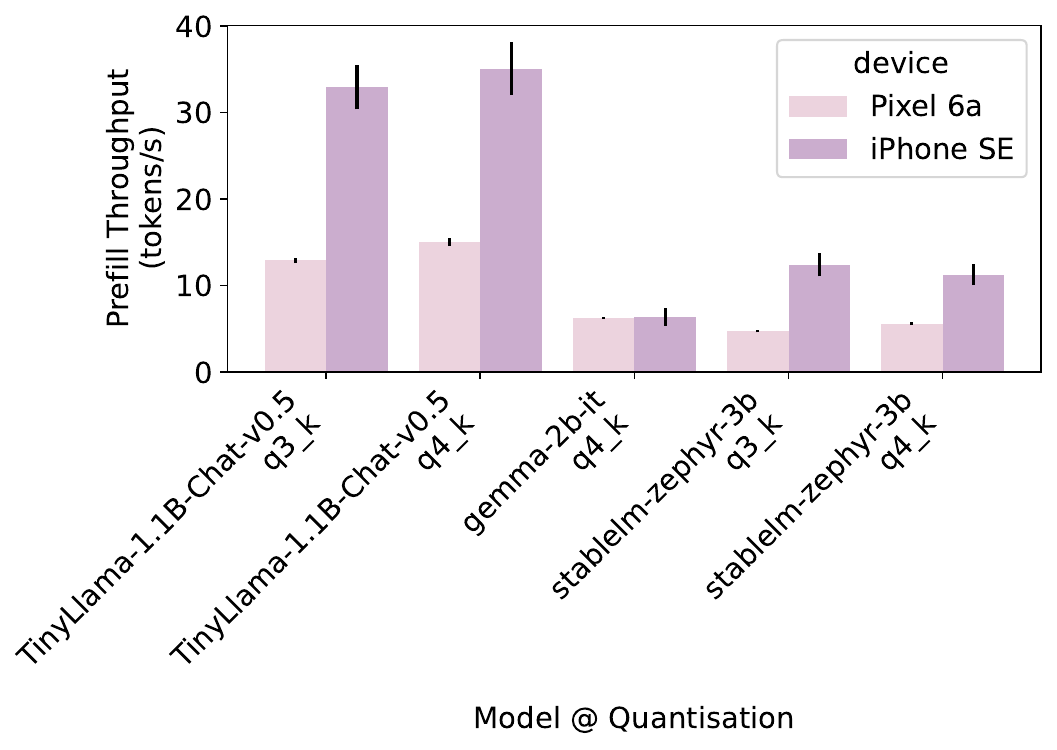}
        \caption{\camready{Prefill throughput for \mbox{llama.cpp} on mid-tier devices (CPU)}}
        \label{fig:prefill_llamacpp_midtier}
    \end{subfigure}
    \begin{subfigure}[t]{0.23\textwidth}
        \centering
        \includegraphics[width=\linewidth,trim={0 1cm 0 0},clip]{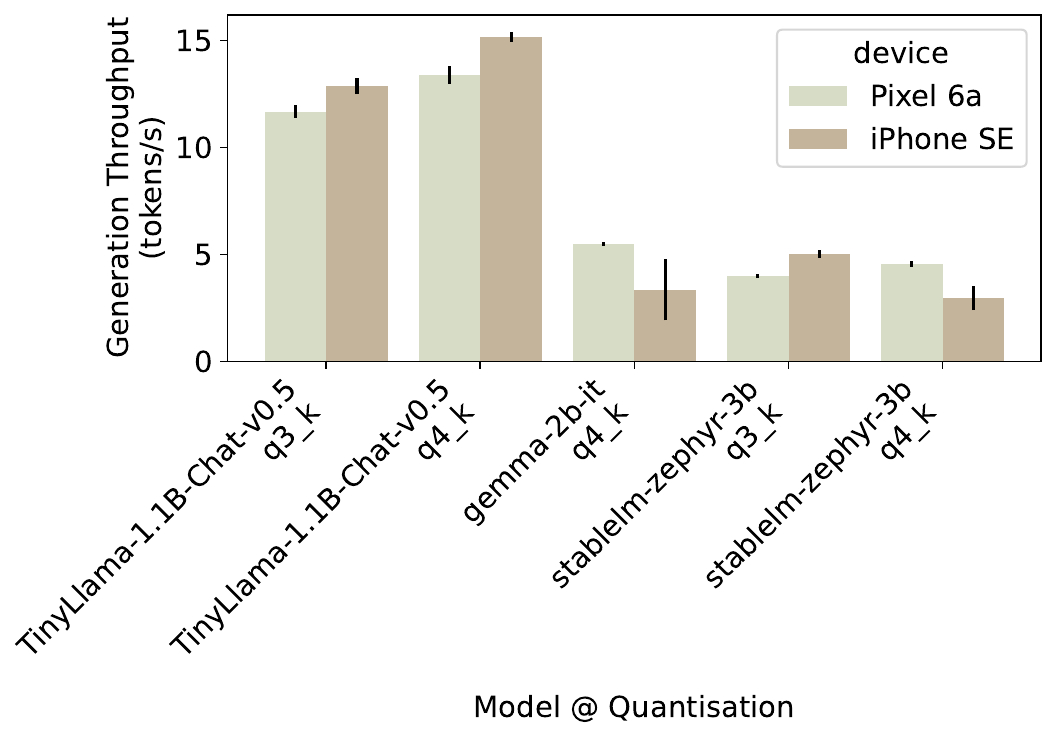}
                \caption{\camready{Generation throughput for \mbox{llama.cpp} on mid-tier devices (CPU)}}
        \label{fig:generate_llamacpp_midtier}
    \end{subfigure}
    \vspace{-0.4cm}
    \caption{Throughput across frameworks and devices}
    \label{fig:prefill_generate_throughputs}
\end{figure}

\begin{figure*}[t]
    \centering
    \begin{subfigure}[t]{0.32\textwidth}
        \centering
        \includegraphics[width=\linewidth]{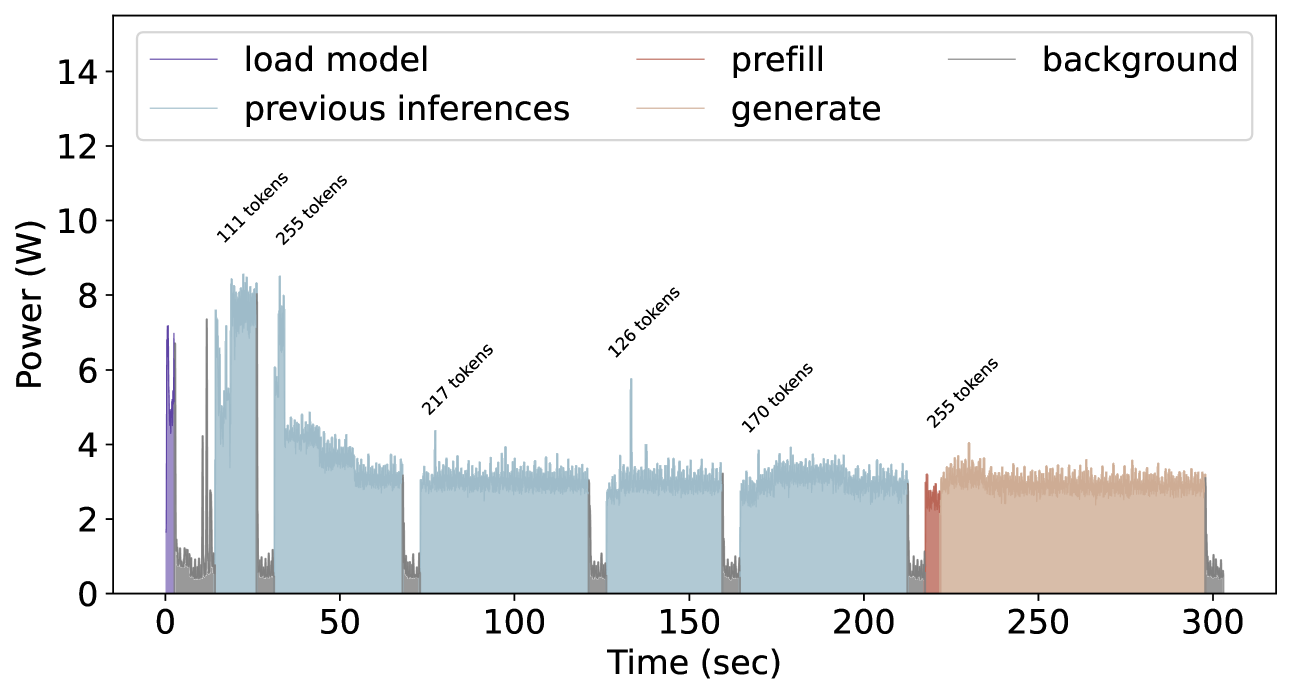}
        \vspace{-0.6cm}
        \caption{MLC-LLM on Galaxy S23 (GPU)}
        \label{fig:s23_mlc_timeline}
    \end{subfigure}
    \begin{subfigure}[t]{0.32\textwidth}
        \centering
        \includegraphics[width=\linewidth]{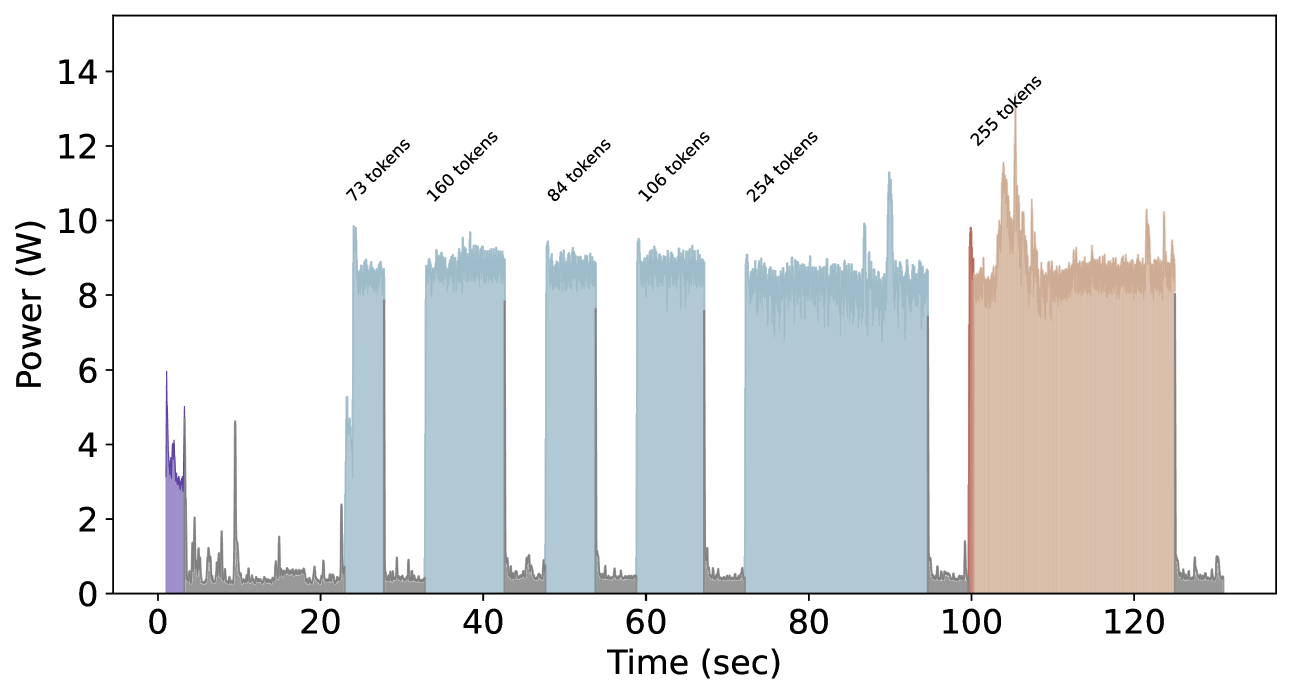}
        \vspace{-0.6cm}
        \caption{MLC-LLM on iPhone 14 Pro (GPU)}
        \label{fig:iphone14_mlc_timeline}
    \end{subfigure}
    \begin{subfigure}[t]{0.32\textwidth}
        \centering
        \includegraphics[width=\linewidth]{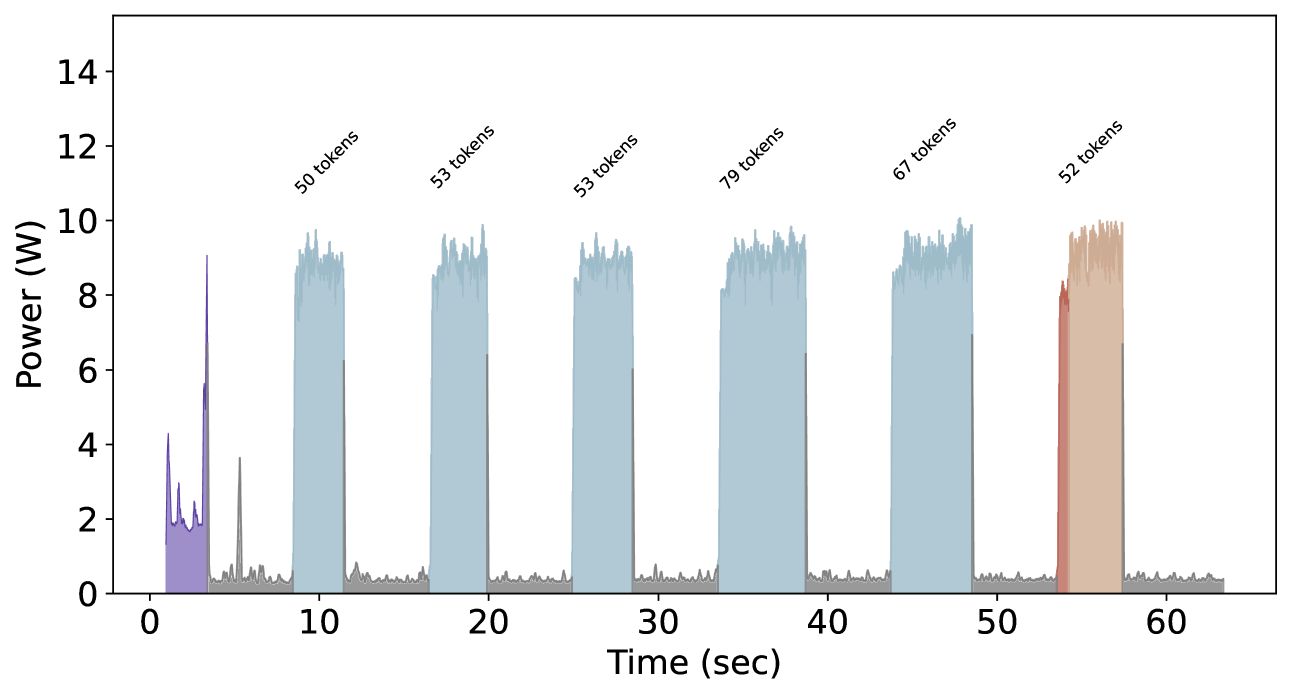}
        \vspace{-0.6cm}
        \caption{LLMFarm on iPhone 14 Pro (GPU)}
        \label{fig:iphone14_llmfarm_timeline}
    \end{subfigure}
    \vspace{-0.4cm}
    \caption{LLM execution timeline of Zephyr-3B (4-bit quantized) across devices and frameworks. We use a moving average of 500 points for smoothing the timeline. We annotate the number of generated tokens per inference.}
    \label{fig:llm_timelines}
\end{figure*}

\noindent
\textbf{Computational throughput.} First, we show the prefill and generation throughput of various models when used in a conversational setting. We divide our results per device tier and illustrate the average throughput (in tokens/sec) per framework in Fig~\ref{fig:prefill_generate_throughputs}. Generally, we witness much higher prefill vs. generation throughput, which can be largely attributed to the usage of KV-cache~\cite{pope2023efficiently} when encoding a sequence of tokens and the compute vs. memory boundedness of the workload~\cite{mediapipe}. Moreover, MLC-LLM generally offered higher performance to llama.cpp, but at the cost of model portability (models need to be compiled per platform). \camready{Operator fusion and TVM-based optimization play a significant role towards this result, with generation throughput difference of +4\% on average for GPU execution (+28\% vs llama.cpp CPU) and up to 3.53$\times$ higher.} Notable exceptions included TinyLlama across targets and Gemma on S23.
We also noticed that 4-bit quantized models performed better than their 3-bit variants, offering 24.77$\%$ higher throughput on average. We attribute this to the effects of dequantization and better cache alignment during execution. However, there is a trade-off with memory consumption, which made certain models to run out-of-memory during runtime, especially on phones with smaller RAM sizes.
Last, the Metal-accelerated iPhones seem to be offering higher throughput compared to the OpenCL-accelerated Android phones for the case of MLC, with 78.93\% higher generation throughput on average. \camready{Even in the case of CPU runtime on llama.cpp, iPhones generally performed faster, but less efficiently. We can attribute this to the higher thread count that llama.cpp allowed on iOS without crashing the application.}
Last, our hypothesis for the relatively high variance of the results is the variable context and generation length as well as potential Dynamic Voltage and Frequency Scaling (DVFS) (more in commentary of Fig.~\ref{fig:llm_timelines}). \camready{We provide performance and discharge rates for GPU execution for llama.cpp on iOS devices in Appendix~\ref{app:gpu_llmfarm}.}

\begin{figure}[t]
    \centering
    \begin{subfigure}[t]{0.23\textwidth}
        \centering
        \includegraphics[width=\linewidth,trim={0 1cm 0 0},clip]{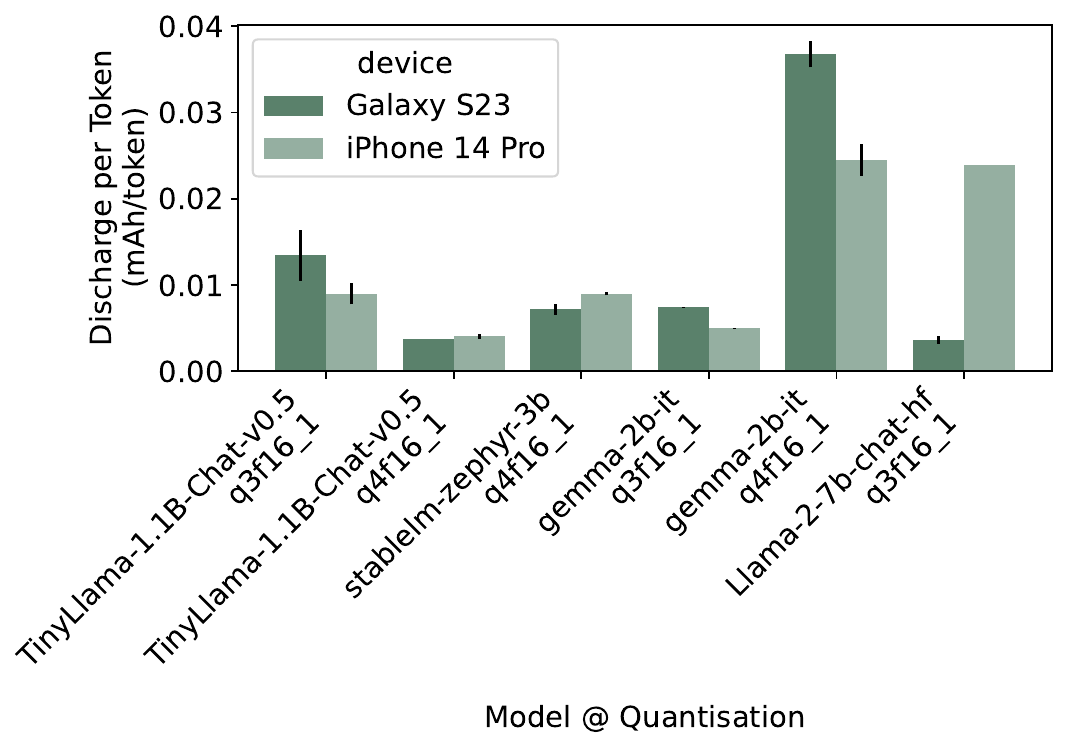}
        \caption{MLC-LLM on high-tier devices (GPU)}
        \label{fig:discharge_per_token_mlc_hightier}
    \end{subfigure}
    \begin{subfigure}[t]{0.23\textwidth}
        \centering
        \includegraphics[width=\linewidth,trim={0 1cm 0 0},clip]{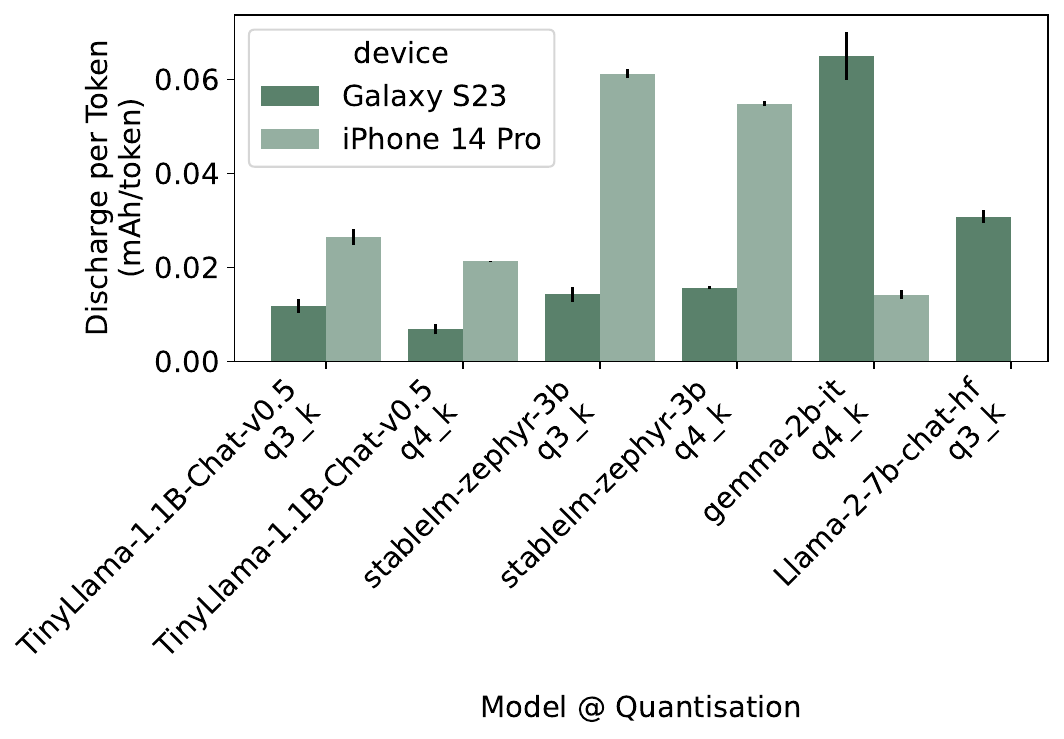}
        \caption{\camready{llama.cpp on high-tier devices (CPU)}}
        \label{fig:discharge_per_token_llamacpp_hightier}
    \end{subfigure}
    \begin{subfigure}[t]{0.23\textwidth}
        \centering
        \includegraphics[width=\linewidth,trim={0 1cm 0 0},clip]{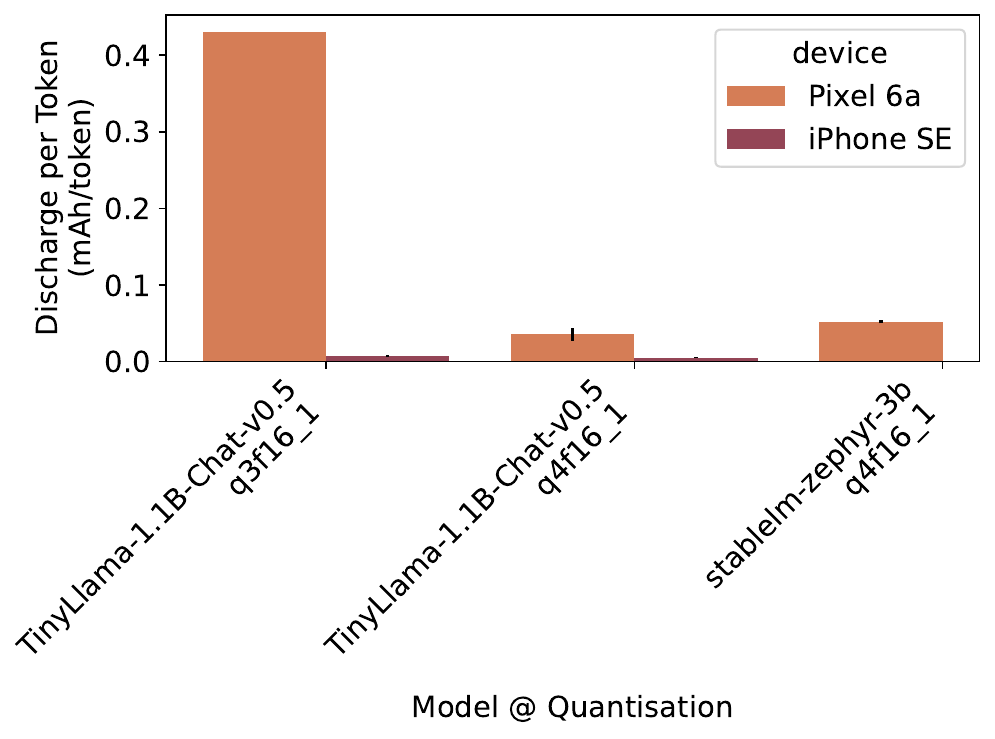}
        \caption{MLC-LLM on mid-tier devices (GPU)}
        \label{fig:discharge_per_token_mlc_lowtier}
    \end{subfigure}
    \begin{subfigure}[t]{0.23\textwidth}
        \centering
        \includegraphics[width=\linewidth,trim={0 1cm 0 0},clip]{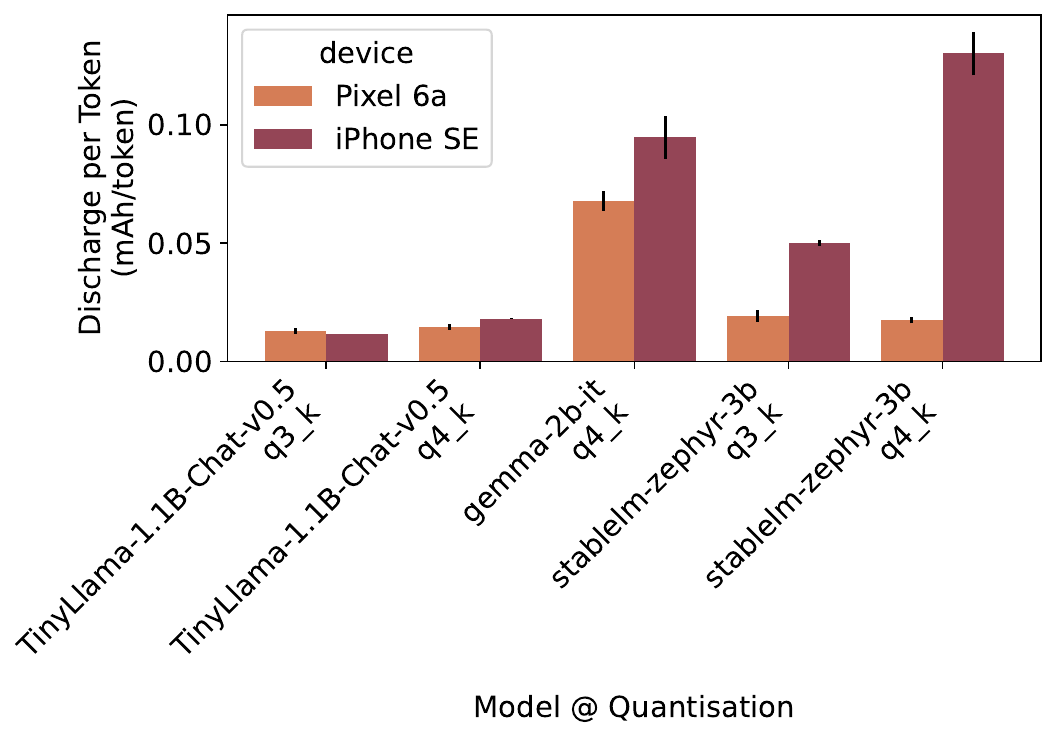}
        \caption{\camready{llama.cpp on mid-tier devices (CPU)}}
        \label{fig:discharge_per_token_llamacpp_lowtier}
    \end{subfigure}
    \vspace{-0.4cm}
    \caption{Discharge per token across frameworks and devices. Missing bars indicate unsuccessful runs due to OOM errors or too long runtime ($>$1hr per conversation).}
    \label{fig:discharge_per_token}
\end{figure}

\noindent
\textbf{Energy efficiency.} Next, we take the same set of models and illustrate the energy discharge (in mAh) per token generated across devices and frameworks in Fig.~\ref{fig:discharge_per_token}.
Overall, we noticed that the trend of larger networks (in terms of parameter size) offering larger discharge rates across devices and frameworks. This is expected as DRAM utilization and memory copies into the SoC registers consume significant energy~\cite{patterson2022carbon}. Notable exceptions to this rule were TinyLlama (3-bit) and Gemma (4-bit), which we aim to investigate with help from upstream maintainers. Last, the CPU execution of llama.cpp offered overall lower energy efficiency, but this could also be attributed to the latency of running inference compared to when using GPU acceleration.

\noindent
\textbf{Power timeline.}
Next, we zoom into the runtime of our experiments and show the execution timeline of Zephyr-3B (4-bit quantized) running six prompts across devices (iPhone 14 Pro and Galaxy S23) and frameworks (MLC-LLM and LLMFarm). During execution, we have traced specific events of interest, that we annotate on Fig.~\ref{fig:llm_timelines}, which depicts the power draw (in Watts) of the device during inference. 
First off, we noticed from the beginning that iPhones tend to boost their power draw very high, reaching a maximum of 13.8W of sustained (averaged) power draw and an instantaneous maximum of over 18W. The equivalent wattage from the Galaxy device only reached an instantaneous maximum of 14W with sustained power draw below 8.5W. 
At the given power draw, the overall power consumption during inference was 11.54, 10.43, 2.42 mWh (normalized per token: 0.21, 0.20, 0.16 mWh/token) for S23 and iPhone 14 Pro on MLCChat and LLMFarm, respectively. At that pace, each device could run 542.78, 490.05 and 590.93 prompts until its battery is depleted, at an average input of 40 tokens and generation length of 135 tokens. Of course, we do not account for simultaneous load and different energy modes applied by the OS here.

Model loading on Zephyr-3B (4-bit) took an average of 2.41{\small$\pm$0.09} sec, during which time the device often becomes unresponsive. We annotate with blue color five intermediary conversations and divide the sixth into prefill and generate events. In between runs, we have a sleep of 5 seconds. As also previously discussed, prefill takes only a small part of the inference which is mostly bottlenecked by the memory-bound generation process.
The length of each inference is not only a function of the speed of the device, but also the number of generated tokens and context size. Therefore, we see the time length of each inference varying per device and across devices.
Last, we see that the power drops gradually after an inference has completed, which is signified by the last gray spike per inference. We remind to the reader at this point that we synchronize the clocks of the co-ordinator and client device to avoid time drift issues.

\subsubsection{Quality of Experience (QoE)}
\label{sec:qoe}

\begin{figure}[t]
    \vspace{-0.3cm}
    \centering
    \includegraphics[width=.6\linewidth]{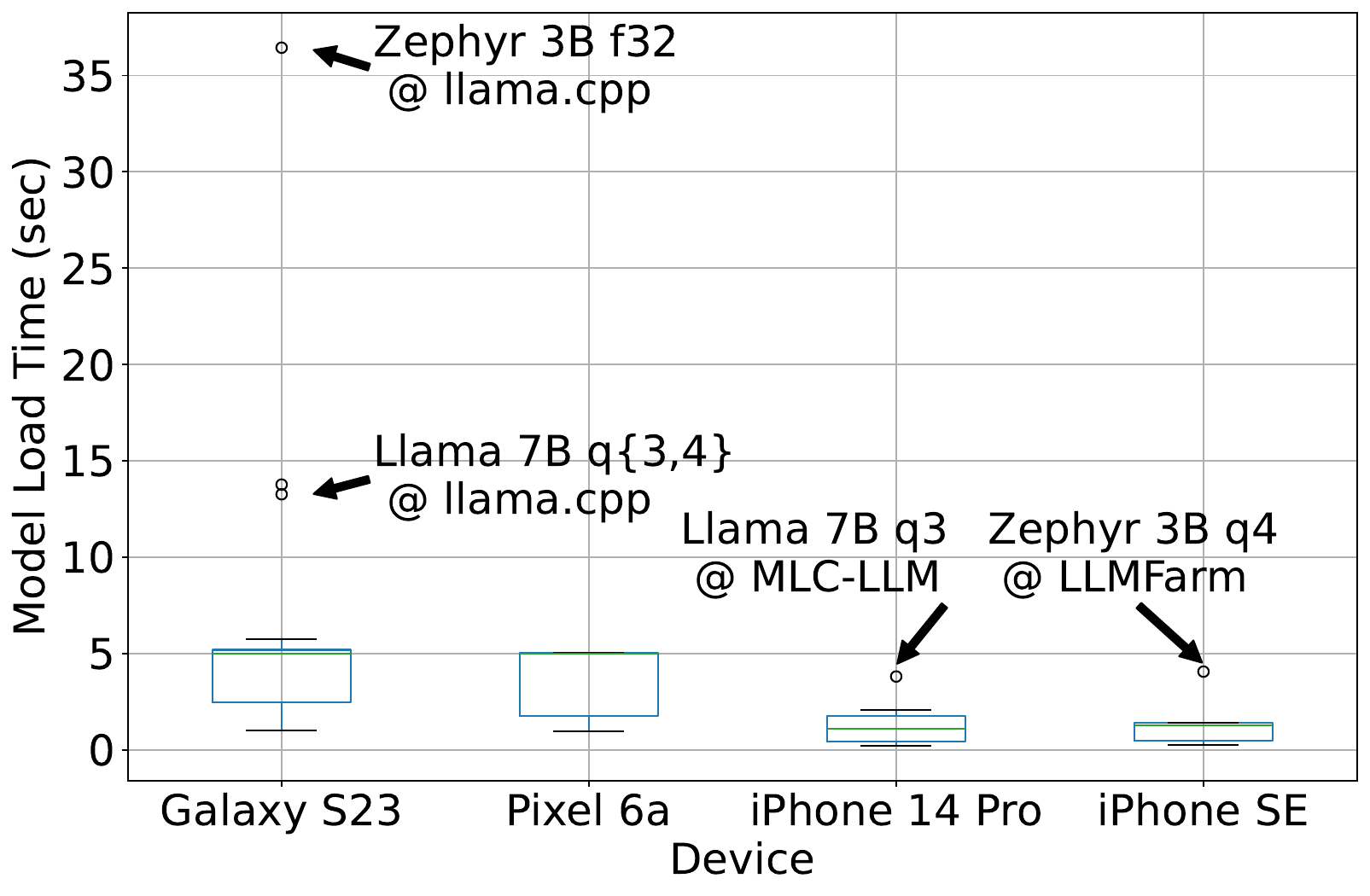}
    \vspace{-0.4cm}
    \caption{Model loading time per device. Each supports different set of models, based on available memory and framework.}
    \label{fig:loading_times}
\end{figure}

In real-world settings, tractability does not imply deployability. What this means is that while a model can run on a device, it can adversely affect the user experience and render the device unstable or unusable. There are largely three dimensions to consider: \textit{i)}~device responsiveness, \textit{ii)}~sustained performance and \textit{iii)}~device temperature. We discuss each of them below: 

\noindent
\textbf{Device responsiveness} 
refers to the general stability and reliability of the device during the runtime of LLM inference. Upon deployment, factors that affected the device responsiveness included long \textit{model loading times} (see purple areas in Fig.~\ref{fig:llm_timelines} and Fig.~\ref{fig:loading_times}) during which the device became largely unresponsive\footnote{Inference impacts overall usability as GPU is also used for GUI rendering.}; \textit{out-of-memory errors} (OOM), which killed the application at arbitrary times; and \textit{device restarts}, which for undefined reasons caused Denial of Service (DoS) by rebooting the device. All these negatively affect the user experience and their frequency of appearance should be minimized. We encountered multiple such events during our benchmarks, which create the need for heterogeneous in-the-wild deployments and parameter selection (e.g.,~model size, quantization precision, prefetching, KV cache size, batch size, context size) based on the available device resources and use-case at hand.

\begin{figure}[t]
    \vspace{-0.3cm}
    \centering
    \begin{subfigure}[t]{0.23\textwidth}
        \centering
        \includegraphics[width=\linewidth]{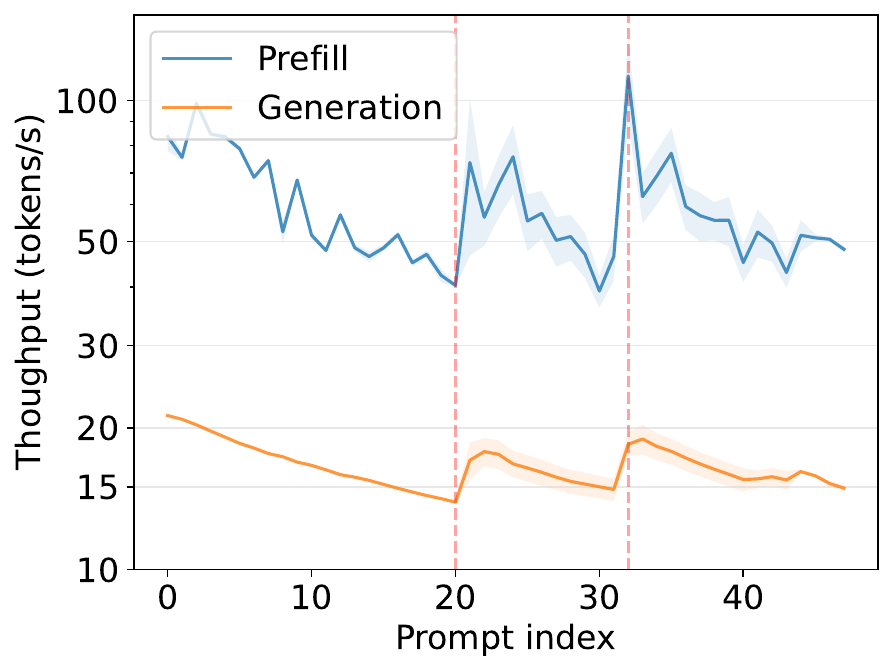}
        \vspace{-0.6cm}
        \caption{iPhone 14 Pro on LLMFarm (GPU)}
        \label{fig:continuous_inference_iphone}
    \end{subfigure}
    \begin{subfigure}[t]{0.23\textwidth}
        \centering
        \includegraphics[width=\linewidth]{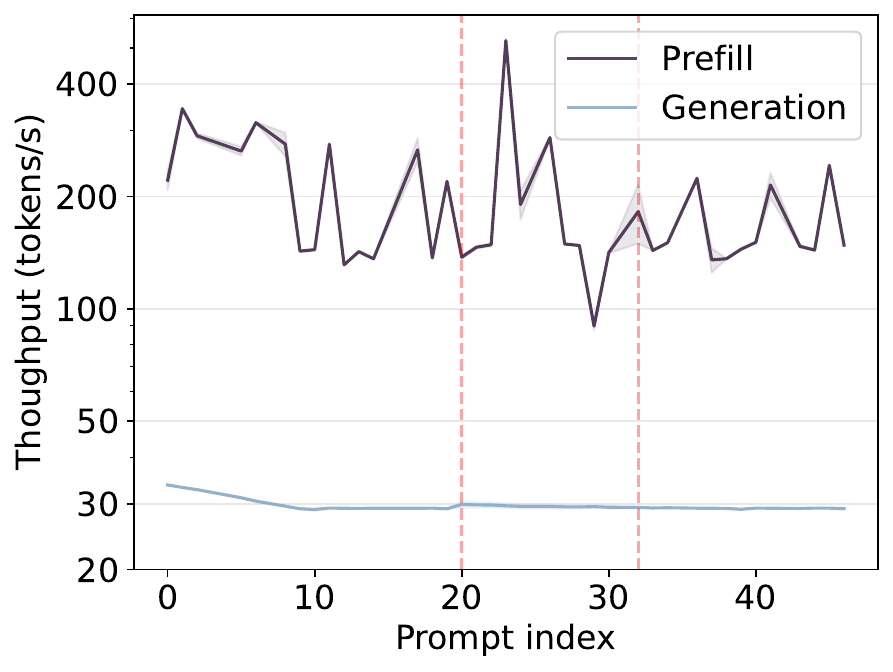}
        \vspace{-0.6cm}
        \caption{Jetson AGX (50W) on llama.cpp}
        \label{fig:continuous_inference_jetson}
    \end{subfigure}
    \vspace{-0.4cm}
    \caption{Continuous inference on mobile and edge devices with Zephyr-3B (4-bit).}
    \label{fig:sustained_inference}
\end{figure}

\noindent
\textbf{Sustained performance}
refers to the device's ability to offer the same performance throughout the runtime of multiple inference requests. There are multiple reasons why this may not be stable, including DVFS, thermal throttling, different power profiles, low battery level and simultaneous workloads, among others.
At this stage, we assume that our LLM is the main workload running on the device, although it has been reported that multiple DNNs reside on smartphones nowadays~\cite{laskaridis2022future}.
To quantify how, we took Zephyr-3B (4-bit) on iPhone 14 Pro and ran continuous inference over 50 prompts to check where throughput starts degrading. We repeated the experiment three times and measure the variation across runs. Results are depicted on Fig.~\ref{fig:continuous_inference_iphone}. We experience straightaway performance dropping with two bumps happening on the 20th and 32nd prompts (on average, annotated in red). Our hypothesis is that the device enters different energy and DVFS modes at these stages, with higher variation signifying that the point at which this happens is not fixed in time. The performance on Jetson AGX (50W) was much smoother (Fig.~\ref{fig:continuous_inference_jetson}), as signified by the straight line in the generation throughput. The initial higher generation throughput can be attributed to the context not being filled.

\begin{figure}[t]
    \vspace{-0.3cm}
    \centering
    \begin{subfigure}[t]{0.48\textwidth}
        \centering
        \includegraphics[width=0.4\linewidth]{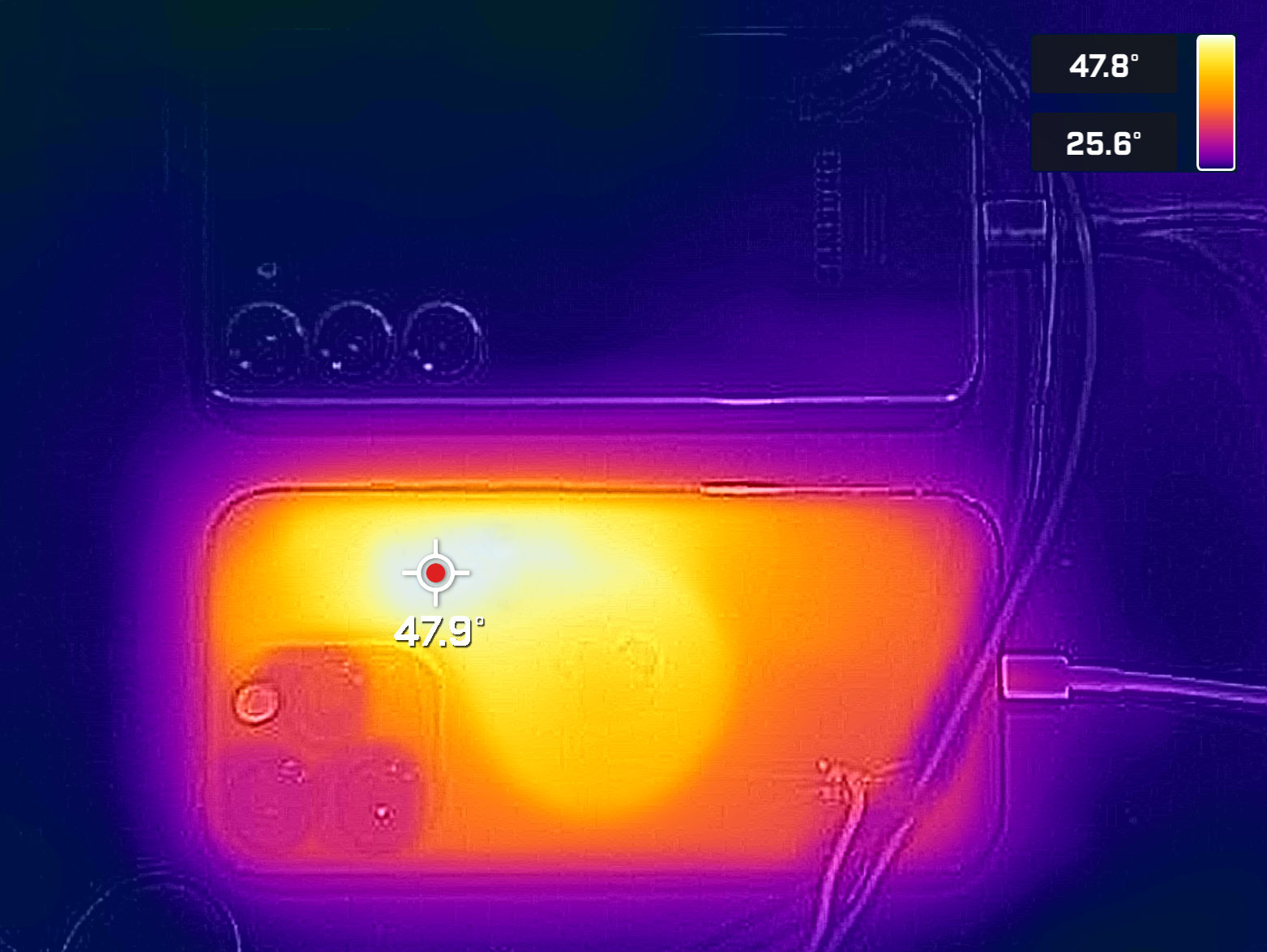}
        \vspace{-0.1cm}
        \caption{Temperature after a full conversation on Zephyr-3B (4-bit) on MLC-LLM}
        \label{fig:iphone14_mlc_temperature}
    \end{subfigure}
    \begin{subfigure}[t]{0.45\textwidth}
        \centering
        \includegraphics[width=.95\linewidth]{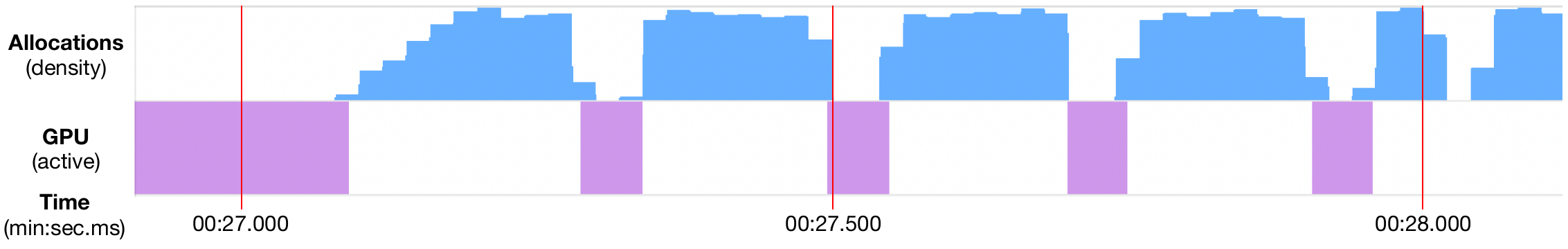}
        \vspace{-0.1cm}
        \caption{Memory trace when running Zephyr-3B (4-bit) on LLMFarm (GPU)}
        \label{fig:iphone14_memory}
    \end{subfigure}
    \vspace{-0.4cm}
    \caption{Thermal and memory behavior on iPhone 14 Pro}
    \vspace{0.2cm}
\end{figure}

\noindent
\textbf{Temperature} 
is yet another parameter that we briefly touched upon in the previous paragraph. Temperature does not only affect device performance, but also user comfort~\cite{10.1145/1978942.1979316}. Devices nowadays come in various forms, but mostly remain passively cooled. Therefore, heat dissipation is mainly facilitated by the use of specific materials and heat management is governed by the OS. The power draw that was witnessed in Fig.~\ref{fig:iphone14_mlc_timeline} did cause temperatures to rise to uncomfortable levels, reaching 47.9\textdegree C as depicted in Fig.~\ref{fig:iphone14_mlc_temperature}.

\begin{figure*}[t]
    \vspace{-0.3cm}
    \centering
    \begin{subfigure}[t]{0.3\textwidth}
        \centering
        \includegraphics[width=.9\linewidth]{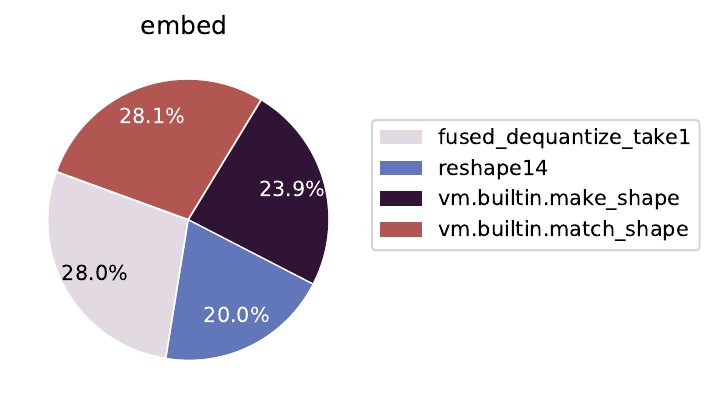}
        \vspace{-0.45cm}
        \caption{Embed}
        \label{fig:per_op_embed}
    \end{subfigure}
    \begin{subfigure}[t]{0.3\textwidth}
        \centering
        \includegraphics[width=.9\linewidth]{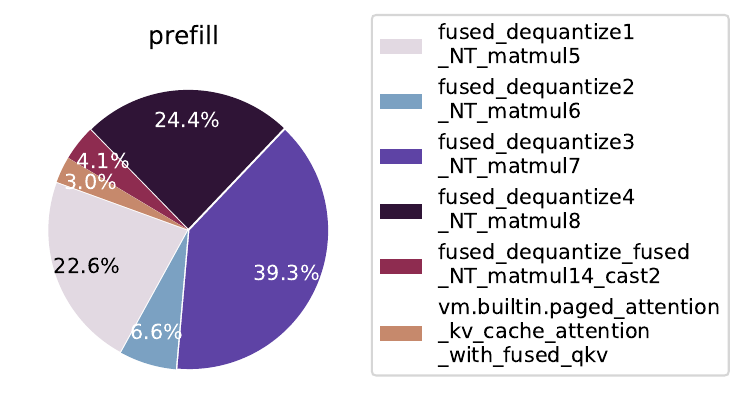}
        \vspace{-0.45cm}
        \caption{Prefill}
        \label{fig:per_op_prefill}
    \end{subfigure}
    \begin{subfigure}[t]{0.3\textwidth}
        \centering
        \includegraphics[width=.9\linewidth]{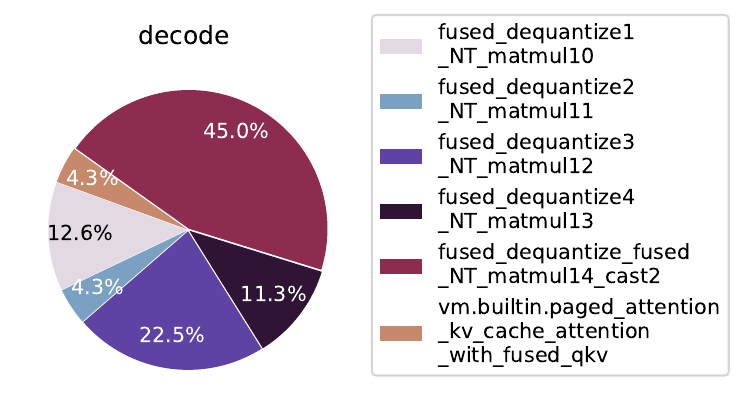}
        \vspace{-0.45cm}
        \caption{Decode}
        \label{fig:per_op_decode}
    \end{subfigure}
    \vspace{-0.4cm}
    \caption{Per-op benchmarks of Llama-7B (3-bit) with MLC-LLM on Samsung Galaxy S23. These are operations generated by the TVM compiler. The variants may signify different implementation or hyperparameters tuned for performance.}
    \label{fig:per_op_benchmarks}
\end{figure*}

\vspace{-0.2cm}
\subsection{Micro Experiments \& Bottlenecks}
\label{sec:microbenchmarks}

In this section, we investigate deeper into on-device LLM inference, and its system bottlenecks. First, we fix the prefill and maximum generation tokens to a fixed number of 256 and remove the \texttt{<EOS>} token for stopping the sequence, leading to a more deterministic execution. In Sec.~\ref{sec:per-op-benchmarking}, we trace the low-level operations during different stages of inference. Next, in Sec.~\ref{sec:bottlenecks}, we use profiling tools to inspect the compute and memory behavior during inference.

\subsubsection{ML Operations}
\label{sec:per-op-benchmarking}

We start by introspecting Llama-7B (3-bit) on Android. We compile a custom version of TVM and MLC-LLM where we enable the \texttt{vm\_profiler} in the backend and report kernel runtimes per operator of interest. In this section, we only measure per kernel latency, as the end-to-end latency is heavily impacted by the use of the profiler. Results are shown in Fig.~\ref{fig:per_op_benchmarks} for the \texttt{prefill}, \texttt{embed} and \texttt{decode} operations. We see that most of the execution is taken up by de-quantize and matrix multiplication fused operations for the \texttt{prefill} and \texttt{decode} operations, taking up 97\% and 95.7\% of the total runtime, respectively. We hypothesize that the dequantization operation is also why 3-bit quantized networks may have performed worse than their 4-bit counterparts, as we discussed in Sec.~\ref{sec:on-device-runtime}. On the contrary, the embed operation seems mostly to be doing tensor conversion and retrieval operations. Since the generation process is mostly bottlenecked by the \texttt{decode} operation (evident also in Fig.~\ref{fig:prefill_generate_throughputs} and \ref{fig:sustained_inference}), we proceed to investigate the real system bottleneck during execution via profiling. Due to lack of GPU tracing via the Android GPU Inspector on Galaxy S23, we apply the analysis on the iPhone 14 Pro.

\vspace{-0.2cm}
\subsubsection{Memory Usage and Bottlenecks}
\label{sec:bottlenecks}

It is known that LLM execution is bottlenecked by the memory bandwidth requirements during generation~\cite{kwon2023efficient,dao2022flashattention,dao2023flashattention}. Effectively, inference waits for the model state and activations to be expensively transferred from main to the on-chip memory, with little reuse due to the small batch sizes and autoregressive causal generation nature of the workload. 
Our analysis corroborates this on the mobile side, by what is shown in the memory profiling of Fig.~\ref{fig:iphone14_memory}, where we depict the memory allocations and GPU computation happening effectively one after the other. While GPU memory gets allocated, GPU compute effectively stalls, waiting for data to process. This was measured through \texttt{xctrace} and visualized with Apple Instruments application.

\vspace{-0.2cm}
\subsection{Impact of Quantization}
\label{sec:quantization_acc}

A prominent method for reducing the memory traffic between main and on-chip memory is to decrease the precision of the weights and activations of the Neural Network~\cite{frantar2022gptq, xiao2023smoothquant,lin2023awq}. However, this often comes at the expense of model accuracy, especially at sub 4-bit weight precision. Moreover, the hardware needs to support operations at these precisions, to avoid dequantization before computation.

By leveraging the supported quantization schemes in the two LLM frameworks \tool supports (Tab.~\ref{tab:llm-frameworks}), we measure the impact of quantization in various tasks on the pretrained models. {We use pretrained instead of fine-tuned models for this because the latter's fine-tuning and RLHF~\cite{ouyang2022training} alignment can affect the original performance}. A description of the employed quantization schemes is presented in Sec.~\ref{sec:related_work}. We use the benchmark datasets depicted in Tab.~\ref{tab:evaluation_datasets}, which consist of Natural Language Inference (NLI) and Natural Language Generation (NLG) tasks. In the former case, it comprises multiple choice questions, and the most likely answer -- expressed by cumulative log likelihood of the model's output -- is selected and matched against the correct label. In the latter case, the model's output is evaluated against template answers over BLEURT~\cite{sellam2020bleurt} score.

\begin{figure*}[t]
    \vspace{-0.5cm}
    \begin{subfigure}[t]{0.2\textwidth}
        \centering
        \includegraphics[width=\linewidth]{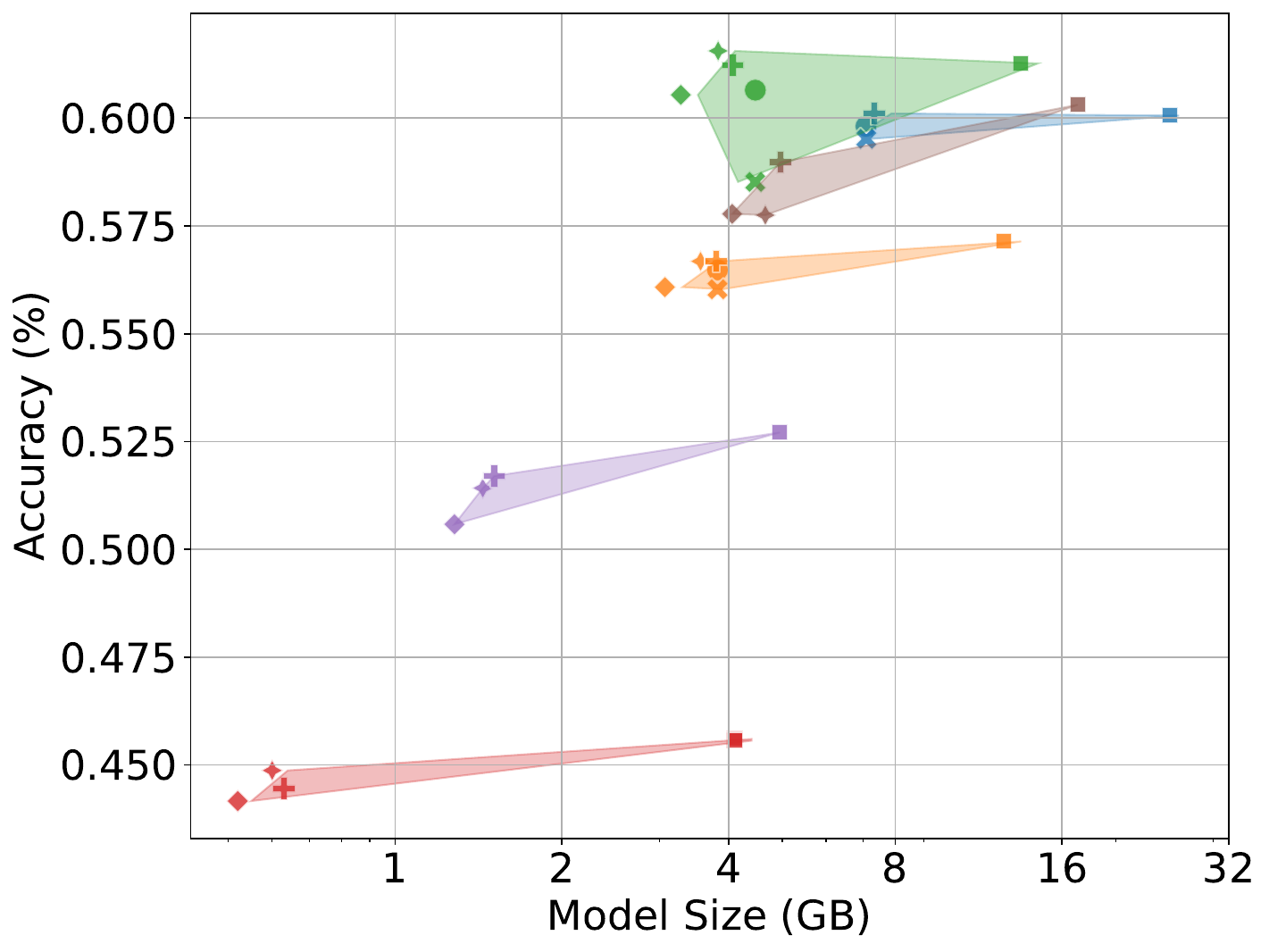}
        \vspace{-0.5cm}
        \caption{HellaSwag} 
        \label{fig:hellaswag-quant-performance}
    \end{subfigure}
    \begin{subfigure}[t]{0.2\textwidth}
        \centering
        \includegraphics[width=\linewidth]{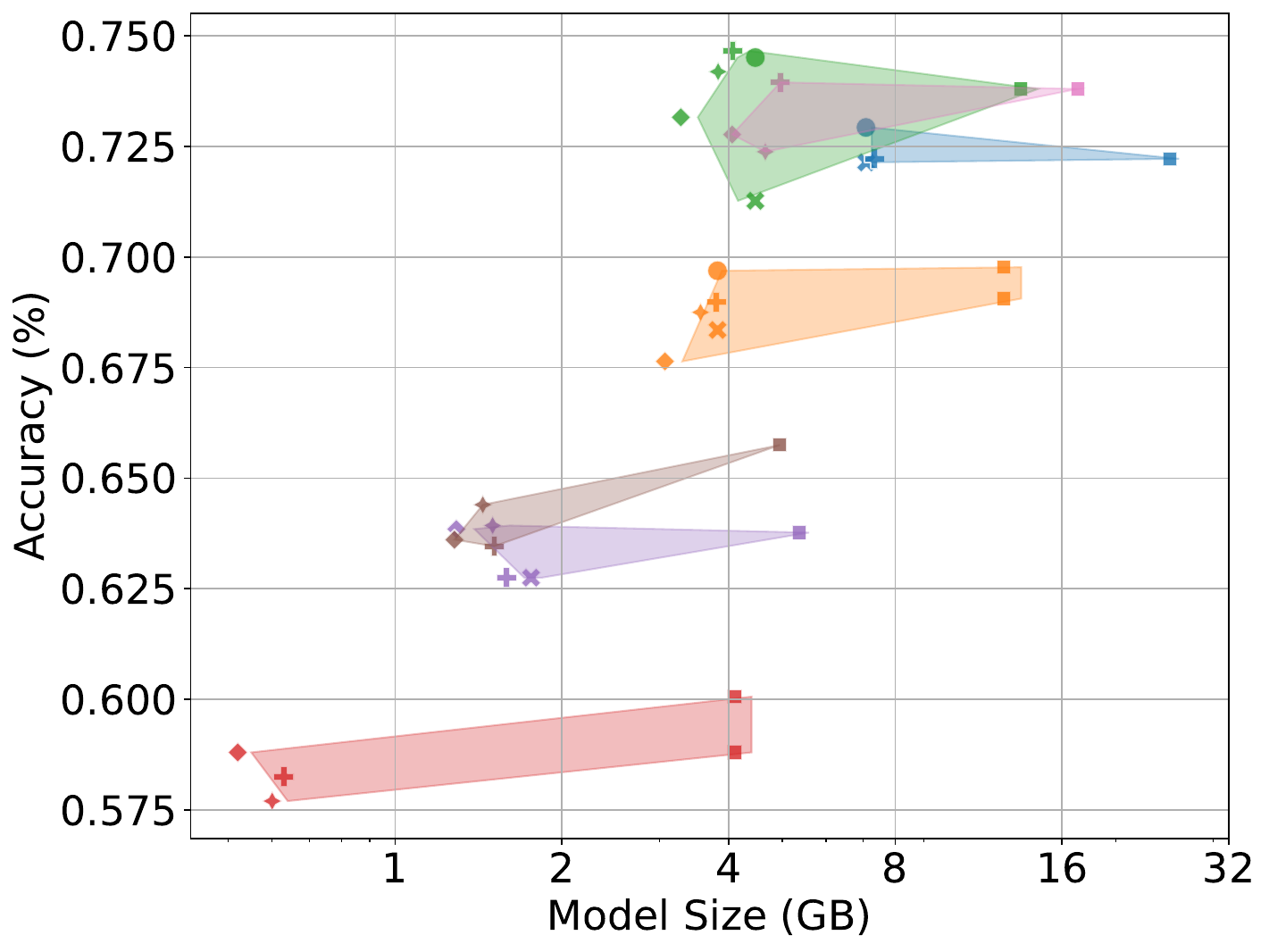}
        \vspace{-0.5cm}
        \caption{Winogrande}
        \label{fig:winogrande-quant-performance}
    \end{subfigure}
    \begin{subfigure}[t]{0.2\textwidth}
        \centering
        \includegraphics[width=\linewidth]{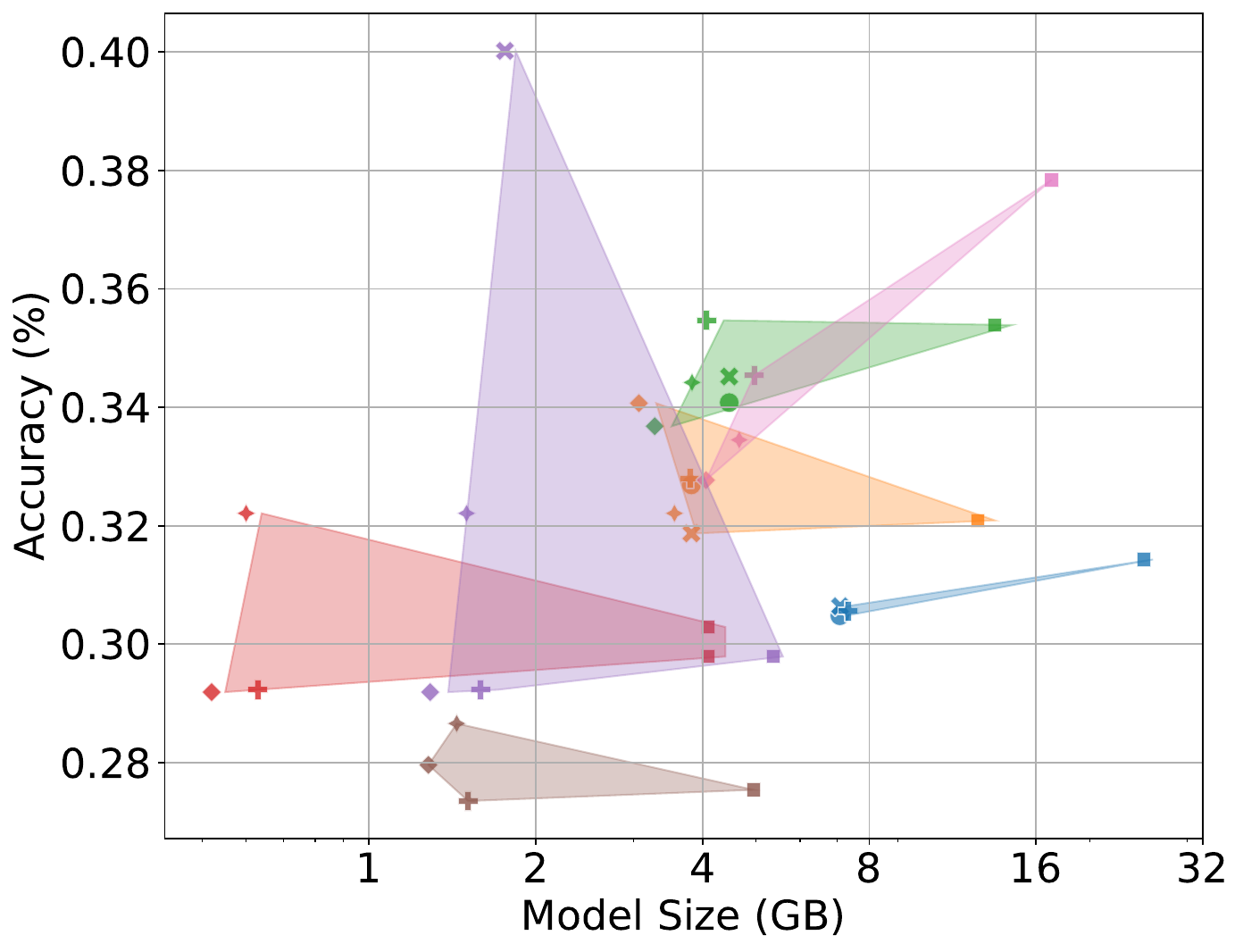}
        \vspace{-0.5cm}
        \caption{TruthfulQA}
        \label{fig:truthfulqa-quant-performance}
    \end{subfigure}
    \begin{subfigure}[t]{0.2\textwidth}
        \centering
        \includegraphics[width=\linewidth]{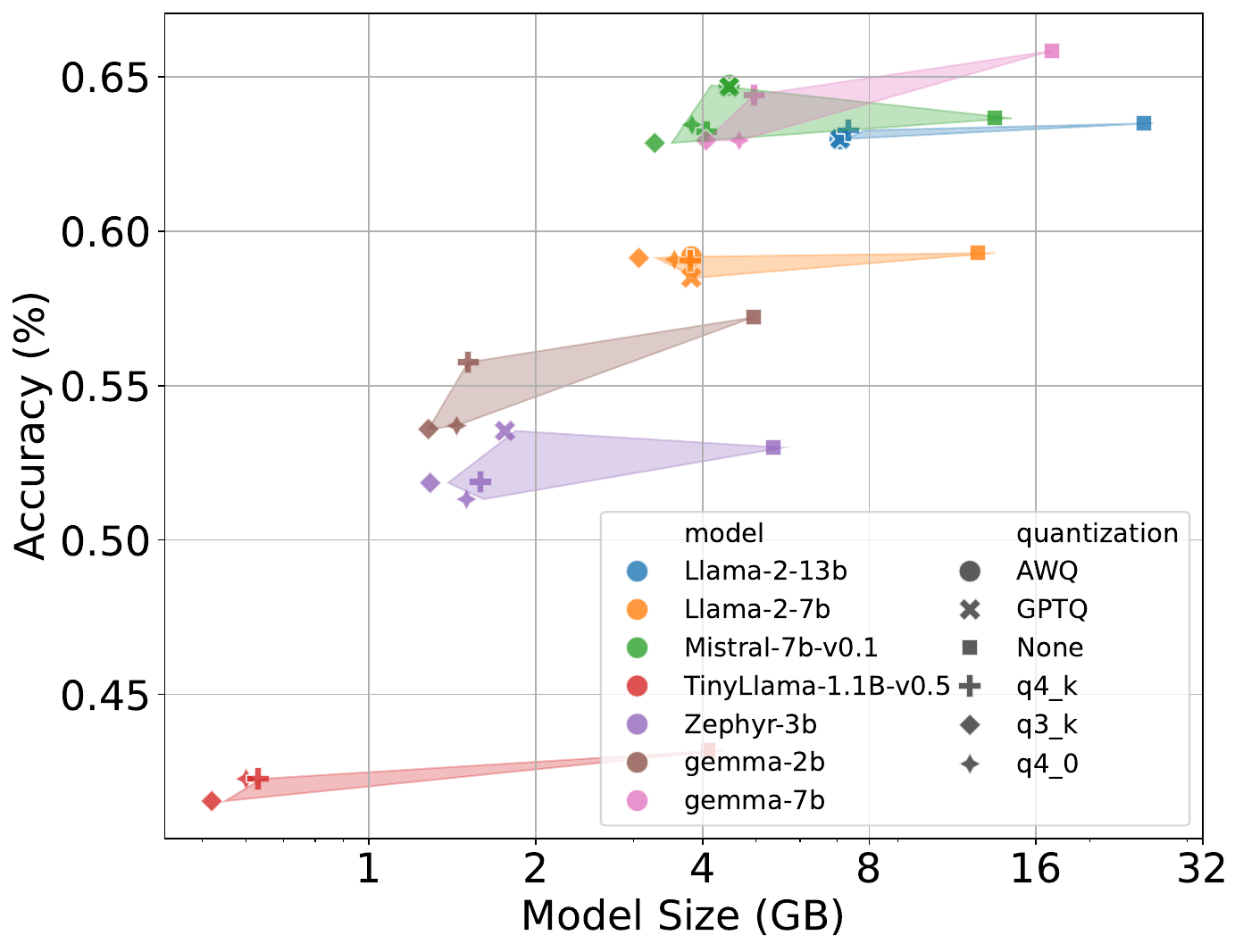}
        \vspace{-0.5cm}
        \caption{ARC}
        \label{fig:arce-quant-performance}
    \end{subfigure}
    \vspace{-0.4cm}
    \caption{Model size vs. accuracy for different models, quantization schemes and precisions.}
    \vspace{-0.4cm}
    \label{fig:quant-performance}
\end{figure*}

Results are depicted in Fig.~\ref{fig:quant-performance} across datasets and models. 
From the data we can see that the most evident performance difference comes from the \textit{model architecture} and \textit{parameter size}, and this performance difference persists across datasets. In terms of quantization schemes, it is obvious that bitwidth is correlated to model size, but also to accuracy, i.e.,~lower bitwidth means higher error rate. This was very evident in our qualitative evaluations, where some smaller models ($\leq$3B parameters) were unusable with 3-bit precision, mostly hallucinating or plainly repeating the prompt. On the other hand, there was no single quantization scheme that performed uniformly better across the board. For larger models ($\geq$7B parameters), AWQ~\cite{lin2023awq} and GPTQ~\cite{frantar2022gptq} performed slightly better, at the expense of elevated model sizes.

\subsection{Runtime at the Edge}
\label{sec:jetson_evaluation}

\begin{table}[t]
\vspace{-0.2cm}
\centering
\caption{Evaluation datasets description}
\vspace{-0.4cm}
\label{tab:evaluation_datasets}
\begin{adjustbox}{width=.95\linewidth}
\begin{tabular}{l p{2.1cm} p{1cm} p{6.5cm}}
\toprule
\textbf{Dataset} & \textbf{Task} & \textbf{Size} & \textbf{Description} \\
\midrule
\textbf{HellaSwag}~\cite{zellers2019hellaswag} & Common-sense NLI & 70k &  Given an event description, select the most likely continuation.\\
\textbf{Winogrande}~\cite{winogrande}  & Common-sense NLI & 44k & Benchmark for common-sense reasoning, designed not to be easily solvable by statistical models and plain word associations. \\
\textbf{ThuthfulQA}~\cite{lin2021truthfulqa} & Knowledge NLG & 817 & Benchmark for measuring truthfulness in a model's generated answers. \\
\textbf{ARC-\{E,C\}}~\cite{chollet2019measure} & Reasoning NLI & 5.2k, 2.6k & Science and language exam questions from a variety of sources. E: Easy; C: Complex \\
\bottomrule
\end{tabular}
\end{adjustbox}
\end{table}

\noindent
\textbf{Offloading.}
Hitherto, we have witnessed that high-end mobile devices with more than 6GB of memory can run a chat LLM at a reasonable rate. However, this comes at the cost of significant battery depletion (see Sec.~\ref{sec:on-device-runtime}), QoE (see Sec.~\ref{sec:qoe}) and end-task accuracy (see Sec.~\ref{sec:quantization_acc}). 
Therefore, we envision that the future of LLM execution can be collaborative and cross-device at the edge~\cite{laskaridis2022future,qualcomm2023whitepaper}. To this direction, we test to see the viability of offloading the DNN execution to a local edge device, which might be a dedicated accelerator (e.g.,~an Edge-AI Hub) or another edge device (e.g.,~a Smart TV or a high-end router). For this reason, we employ two Jetson devices, namely Nano (mid-tier) and AGX (high-tier) to check the viability of this paradigm. We assume prompt offloading happens over a Wi-Fi 6 network (9.6 Gbps) with negligible latency overhead (i.e., streamed text).

\noindent
\textbf{Emulating different devices}
Jetson devices support different energy modes, which configure the number of active cores and their frequency, along with memory frequency to provide different power envelopes. Specifically, Orin AGX supports TDPs of 50W, 30W, and 15W whereas Orin Nano supports 15W and 7W.
Taking advantage of this functionality, we wanted to test how such devices can support LLM execution under different power budgets and their respective performance. This way, we first showcase the offloading viability over different ambient devices, but also give a proxy metric about potential future mobile and edge devices.

\begin{figure}[t]
    \centering
    \begin{subfigure}[t]{0.23\textwidth}
        \centering
        \includegraphics[width=\linewidth,trim={0 1cm 0 0},clip]{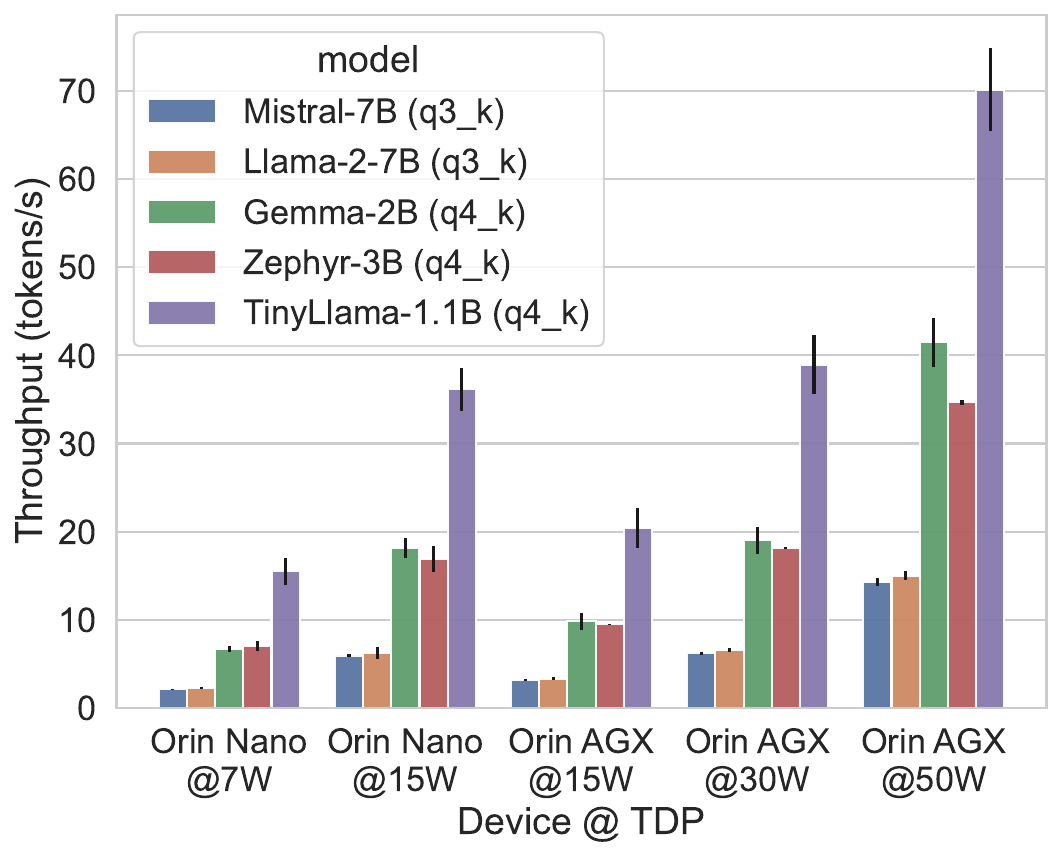}
        \vspace{-0.6cm}
        \caption{Different models token generation throughput on Jetson devices running on llama.cpp.}
        \label{fig:jetson-throughputs}
    \end{subfigure} %
    \begin{subfigure}[t]{0.24\textwidth}
        \centering
        \includegraphics[width=\linewidth,trim={0 1cm 0 0},clip]{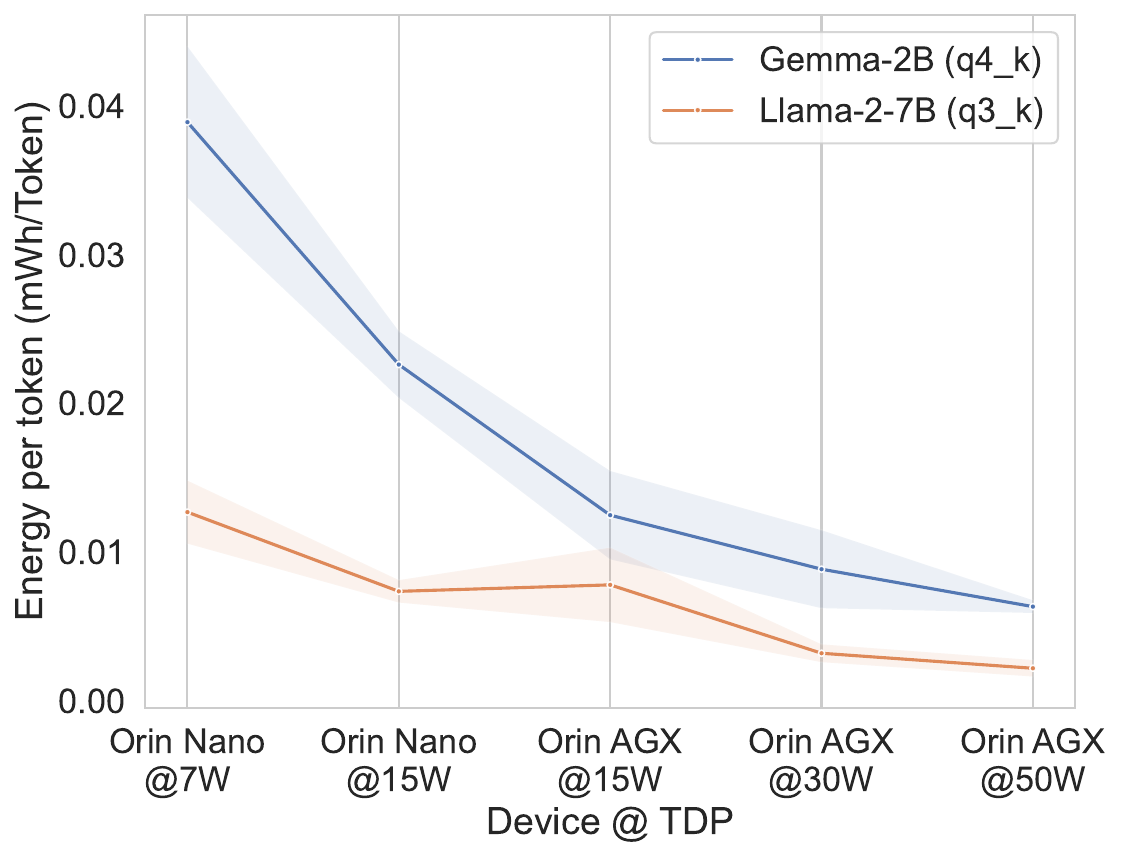}
        \vspace{-0.6cm}
        \caption{Energy consumed per token across energy modes and Jetson devices.}
        \label{fig:jetson-energy}
    \end{subfigure}
    \vspace{-0.4cm}
    \caption{LLM execution on Jetsons across energy modes}
    \label{fig:jetsons}
\end{figure}

Results for both scenarios are presented in Fig.~\ref{fig:jetsons}. Specifically, in Fig.~\ref{fig:jetson-throughputs} we show the generation throughput (in tokens/sec) of various models on different Jetson devices and energy profiles, as run with llama.cpp on CUDA. We see that throughputs largely follow a monotonic trajectory with respect to model size and energy modes, with the notable exception of Orin Nano and Orin AGX at 15W, with the former performing +7.89 tokens/sec 
better on average. Overall, generation throughput is significantly higher than the equivalent mobile runtime, and this runtime can also be sustained for longer periods, as shown in Fig.~\ref{fig:sustained_inference}. Indicatively, for Zephyr-7B (4-bit), the average throughput is 3.3$\times$ and 1.78$\times$ higher, for prefill and generation respectively. \camready{CPU runtimes are also provided in Appendix~\ref{app:jetson_cpu}.}

In Fig.~\ref{fig:jetson-energy}, we quantify the energy efficiency of two models (Llama-7B (3-bit) and Gemma-2B (4-bit)) running across different energy modes. Interestingly and perhaps counter-intuitively, we see that the efficiency is moving the same direction as the device's TDP. 
We believe that frequency scaling of the memory subsystem from the lower power mode adversely affects the workload, making generation even more bottlenecked by the lower memory clock. Effectively, the GPU stalls for longer, waiting for memory I/O. When the power mode allows for higher memory frequency, there are additional efficiency gains and higher utilization.

\vspace{-0.2cm}
\section{Discussion \& Limitations}
\label{sec:discussion}

\textbf{Summary of results.} So far, we visited the performance and energy consumption characteristics of running LLMs on mobile and edge devices. We measured the throughput and energy efficiency of various models and showed that smaller quantized models can run sufficiently well on device at the cost of increased power consumption.
Moreover, we studied the device behavior during model loading and sustained inference, along with the power variability during a conversation, witnessing high peaks and apparent consequences in user QoE.
Last, we dove into the specific operator runtime and memory bottlenecks during execution and showed the memory-bound nature of generation. Recognizing that quantization is one of the main ways to drop the memory requirements, we measured the accuracy impact on various tasks, which was non-negligible in sub 4-bit precisions. 
Drawing from these results, we discuss their impact in LLM deployment and how they can shape future research avenues.

\noindent
\textbf{Hardware/Software advances} While the area of generative AI has seen great acceleration the past years, so have the associated workloads. As an area of active research and industrial interest, new algorithmic methods~\cite{dao2023flashattention,chen2023accelerating,gu2023mamba} and hardware~\cite{fan2022adaptable,luo2023calabash} can provide non-linear scaling in how the current workloads run. Therefore, not only can current models be deployed more efficiently, but also larger models can be trained and deployed, leading to smarter models~\cite{bubeck2023sparks,schaeffer2024emergent}.

\noindent
\textbf{Multimodality \& emergent abilities.} In terms of capabilities, the ability of models to deal with multi-modal inputs and outputs become of great value~\cite{liu2023visual,radford2021learning,mckinzie2024mm1},
effectively giving assistants an extra sense. However, their overhead for deployment is non-negligible, especially on embedded hardware like smart glasses or robotics. Therefore, on-device deployment of such models emerges as an area of interest.

\noindent
\textbf{New use-cases.} This paper is the first step towards enabling use-cases at the edge, offering metrics that can fuel algorithmic and edge hardware research, with efficiency, privacy and sustainability in mind. We envision a future where multi-modal and context-aware personalized assistants will be locally conversing with users and have long-term memory with recollection of past interactions~\cite{10.1145/3580305.3599572}. At the same time, users will be able to interact with interfaces in natural language to accomplish tasks~\cite{li-etal-2020-mapping}, without the need to imperatively define the individual steps~\cite{schick2024toolformer,10.1145/3544548.3580895}. Last, we envision this automation expanding to interactions between humans, where individuals would be able to proxy their availability over smart assistants~\cite{callscreening}.

\noindent
\textbf{Organization of edge hardware resources.} Last, in terms of system architecture, we foresee two major avenues of deploying intelligence at the edge. One requires SoC manufacturers to design accelerators explicitly for running LLMs in an energy efficient manner, in a way that does not hurt QoE of concurrent apps or deplete the battery in an unreasonable manner. To this direction, NPUs capable of running matrix-to-matrix multiplications efficiently with larger on-chip cache and memory throughput seems crucial. The future can also be hybrid~\cite{qualcomm2023whitepaper} and hierarchical, with part of the workload being accelerated at the edge or cloud~\cite{xu2023llmcad,laskaridis2022future,laskaridis_spinn}.

\noindent
\textbf{Limitations.} Our study is simply the first attempt towards analyzing the on-device behavior of LLM workloads and hope can make them more accessible to the public. However, our analysis has been limited to chat fine-tuned models of 1-13B parameter size due to their broad availability and popularity. Very lately, sub-billion models have emerged~\cite{thawakar2024mobillama,liu2024mobilellm}, which present their own computational interest in edge settings. Moreover, we analyzed the inference energy at a device-centric level. It is well known, though, that the consumer edge is not as green as state-of-the-art datacenters~\cite{wu2022sustainable}. The global impact of distributing LLM computation has not been considered. Last, we only studied quantization as a way of reducing model footprint. There are various alternatives, briefly introduced Sec.~\ref{sec:related_work}, for further optimizing these workloads. We leave such topics as future work.

\vspace{-0.2cm}
\section{Related Work}
\label{sec:related_work}
\noindent
\textbf{Benchmarking models on device.}
In terms of on-device DNN benchmarking, there has been a rich set of literature in the past for edge and mobile deployment. Indicatively, Ignatov et al.~\cite{Ignatov_2018_ECCV_Workshops} had been one of the first in-the-wild benchmark suites for on-device benchmarking and device ranking across a multitude of downstream tasks and modalities. Embench~\cite{almeida2019embench} quantified the different dynamics of model execution across various mobile, edge and desktop devices. MLPerf~\cite{reddi2020mlperf} is an industry-wide standardized ML benchmark tool.
Another tangential line of work has focused on quantifying the performance of already deployed models in mobile apps, with works~\cite{almeida2019embench,10.1145/3308558.3313591} showcasing a surging trend in the deployment of on-device ML. Nevertheless, the advent of LLMs have pushed the compute requirements for executing such workloads, and thus current most deployments offload inference to the cloud~\cite{mao2017survey}, while on-device deployment remains limited. This phenomenon is hindered by the currently available tools and asks for better on-device measurements so that edge execution of LLMs is faciliated. 
To the best of our knowledge, this is the first study of LLMs on-device performance. Prior work has either focused on training efficiency~\cite{zeus_2023,10.5555/3433701.3433727} or served inference~\cite{kwon2023efficient,aminabadi2022deepspeed} \mbox{in the datacenter.}

\noindent
\textbf{Edge execution of LLMs.}
There have been various lines of work attempting to port LLM computation on-device. Starting with frameworks, llama.cpp~\cite{llama.cpp} and MLC~\cite{mlc-llm} have stood out, offering cross-platform accelerated execution and support for various LLM architectures and device targets. Other open-source frameworks include llama2.c~\cite{llama2.c}, aimed at simplicity without dependencies and tinygrad~\cite{tinygrad}, focused on accelerated execution, but without support quantized mobile execution. Last, TinyChatEngine~\cite{han_tinychatengine} showcased on-device inference with compressed models, but lacks mobile support.
Lately, OS providers have released their own platforms, such as Apple's MLX~\cite{mlx2023} and Google's AICore~\cite{aicore}. The former only \camready{provided support for desktop platforms (M-series SoCs) at the time of writing} and the latter remains closed-source and only deployed on Pixel 8 Pro. Very recently, Google released MediaPipe~\cite{mediapipe} for on-device LLM execution.

\noindent
\textbf{Efficient LLMs.}
As we have shown, these workloads have been largely bottlenecked by the memory size and throughput of the underlying hardware. Therefore, a lot of research has focused on compressing these models to economize on their memory and bandwidth requirements. Various works have proposed quantization~\cite{lin2023awq, frantar2022gptq, xiao2023smoothquant, liu2023llm, dettmers2023spqr, kim2023squeezellm} and sparsification/pruning schemes~\cite{ma2023llmpruner, frantar-sparsegpt}, low-rank methods~\cite{xu2023tensorgpt} and distillation-based solutions~\cite{gu2023knowledge} aimed specifically at LLMs. Orthogonally, one can leverage secondary storage for running LLMs with limited local resources~\cite{flexgen_icml23,alizadeh2023llm}.
The quadratic cost of attention has also been a large scalability issue. Therefore, various techniques try to address this cost, through different attention patterns~\cite{wang2020linformer,beltagy2020longformer,dao2022flashattention,dao2023flashattention}, token skipping~\cite{guan-etal-2022-transkimmer,kim-cho-2021-length,goyal2020power} or alternative architectures~\cite{peng2023rwkv,gu2023mamba}.

Employing multiple models for dropping the overall cost of inference has also been a popular approach, with techniques such as \emph{Mixture-of-Experts}~\cite{frantar2023qmoe,yi2023edgemoe,fedus2022switch} focusing on using subsets of weights based on the input at hand. However, these remain difficult to deploy on device, due to their memory and storage requirements. \emph{Speculative decoding}~\cite{chen2023accelerating,medusa} has been recently introduced as a way of accelerating inference, based on the fact that not every token needs to be generated by a large LLM, but a significantly smaller draft model can be leveraged for quick token generation while the original model operates in a batched fashion. \cite{xu2023llmcad} proposes a distributed such setup for the edge.
For a more complete overview of related work, we divert the reader to~\cite{wan2023efficient, xu2024survey}.

\vspace{-0.2cm}
\section{Conclusion}
\label{sec:conclusion}
In this work, we have made the first step towards quantifying the performance of deploying LLMs at the consumer edge.
We measured the performance, memory, and energy requirements of such workloads across different model sizes and a heterogeneous ecosystem of devices, pinpointing computational, QoE and accuracy bottlenecks. We hope this study will serve as a basis for subsequent algorithmic and hardware breakthroughs that will help the realization of new use-cases and the democratization of LLMs execution in an open but privacy-preserving manner.

\balance
\bibliographystyle{ACM-Reference-Format}
\bibliography{references}

\newpage
\appendix
\onecolumn
\section*{Supplementary Material}
\section{Additional Evaluation Results}
\label{app:extra-runtimes}

\subsection{\camready{GPU Runtime of LLMFarm on iOS Devices}}
\label{app:gpu_llmfarm}

\begin{table}[h]
    \centering
    \caption{\camready{LLMFarm (iOS) GPU performance and discharge rate for various models on mid-tier devices}}
    \label{table:simplified_device_performance}
    \begin{tabular}{l l c p{2cm} p{3.2cm}}
        \toprule
        \textbf{Device} & \textbf{Model} & \textbf{Quantization} & \textbf{Throughput (tokens/sec)} & \textbf{Discharge per Token (mAh/token)} \\
        \multirow{5}{*}{\textbf{iPhone 14 Pro}} & TinyLlama-1.1B & q3\_k &  22.9826{\tiny±1.0557}    & 0.0085{\tiny±0.0010}  \\
        & TinyLlama-1.1B & q4\_k &  24.6531{\tiny±1.1295}    & 0.0079{\tiny±0.0035} \\
        & Gemma-2B &  q3\_k      &  20.7053{\tiny±0.0949}    & 0.0378{\tiny±0.0091} \\
        & Gemma-2B &  q4\_k      &  23.7938{\tiny±0.1190}    & 0.0322{\tiny±0.0101} \\
        & Zephyr-3B & q3\_k      &  10.5899{\tiny±0.2858}    & 0.0130{\tiny±0.0088} \\
        & Zephyr-3B & q4\_k      &  14.8165{\tiny±0.4524}    & 0.0090{\tiny±0.0011} \\
        & Llama2-7B & q3\_k      &   5.9889{\tiny±0.1374}    & 0.0528{\tiny±0.0105} \\
        \midrule       
        \multirow{4}{*}{\textbf{iPhone SE}} & TinyLlama-1.1B & q3\_k &  30.5524{\tiny±1.1284}  & 0.0074{\tiny±0.0002} \\
        & TinyLlama-1.1B & q4\_k &  31.3951{\tiny±1.0936}  & 0.0066{\tiny±0.0001}  \\
        & Gemma-2B &  q3\_k      &  16.6111{\tiny±0.0713}  & 0.0296{\tiny±0.0004} \\
        & Gemma-2B &  q4\_k      &  16.7310{\tiny±0.1718}  & 0.0289{\tiny±0.0007} \\
        & Zephyr-3B & q3\_k      &  13.7265{\tiny±0.2955}  & 0.0263{\tiny±0.0002}  \\
        & Zephyr-3B & q4\_k      &  12.1618{\tiny±1.4816}  & 0.0364{\tiny±0.0027}  \\
        \bottomrule
    \end{tabular}
        
\end{table}

\subsection{\camready{CPU Runtime of llama.cpp on Jetson Devices}}
\label{app:jetson_cpu}

\begin{table}[h]
    \centering
    \caption{\camready{Jetson CPU performance and energy per token for various models and energy profiles.}}
    \label{table:jetson_cpu}
    \begin{tabular}{l l c p{2cm} p{3cm}}
        \toprule
        \textbf{Device} & \textbf{Model} & \textbf{Quantization} & \textbf{Throughput (tokens/sec)} & \textbf{Energy per token (mWh/token)} \\
        \midrule
        \multirow{5}{*}{\textbf{Orin AGX @ 50W}} & TinyLlama-1.1B & q4\_k & 13.3085{\tiny±0.7917} & 0.0015{\tiny±0.0004} \\
        & Gemma-2B & q4\_k   & 6.2280{\tiny±0.1455}   & 0.0063{\tiny±0.0006} \\
        & Zephyr-3B & q4\_k  & 5.4001{\tiny±0.2857}  & 0.0033{\tiny±0.0013} \\
        & Mistral-7B & q4\_k & 2.2248{\tiny±0.0748} & 0.0201{\tiny±0.0033} \\
        & Llama2-7B & q4\_k  & 2.3284{\tiny±0.0875}  & 0.0204{\tiny±0.0035} \\
        \midrule
        \multirow{5}{*}{\textbf{Orin AGX @30W}} & TinyLlama-1.1B & q4\_k & 10.7740{\tiny±0.6574} & 0.0023{\tiny±0.0007} \\
        & Gemma-2B & q4\_k   & 4.8950{\tiny±0.0925} & 0.0092{\tiny±0.0016} \\
        & Zephyr-3B & q4\_k  & 4.2830{\tiny±0.2189} & 0.0067{\tiny±0.0054} \\
        & Mistral-7B & q4\_k & 1.7442{\tiny±0.0524} & 0.0181{\tiny±0.0056} \\
        & Llama2-7B & q4\_k  & 1.8100{\tiny±0.0644} & 0.0102{\tiny±0.0043} \\
        \bottomrule
    \end{tabular}
\end{table}

\end{document}